\documentclass[10pt,twocolumn,letterpaper]{article}

\usepackage{cvpr}              

\usepackage[justification=justified]{caption}
\usepackage{subcaption}%
\usepackage[font=small,skip=0pt]{caption} 
\usepackage{ragged2e}
\usepackage{bm}
\usepackage{algorithm}
\usepackage{algpseudocode}
\usepackage{multirow}
\usepackage{amsmath}
\usepackage{bibunits}

\definecolor{cvprblue}{rgb}{0.21,0.49,0.74}
\usepackage[pagebackref,breaklinks,colorlinks,allcolors=cvprblue]{hyperref}

\def\confName{CVPR}
\def\confYear{2026}

\title{Sampling-Aware Quantization for Diffusion Models}

\author{
 Qian Zeng\textsuperscript{\rm 1},
 Jie Song\textsuperscript{\rm 1}\thanks{Corresponding author},
 Yuanyu Wan\textsuperscript{\rm 1},
 Huiqiong Wang\textsuperscript{\rm 2},
 Mingli Song\textsuperscript{\rm 1,3,4} \\[2mm]
 $^1$School of Software Technology, Zhejiang University \\
 $^2$Ningbo Innovation Center, Zhejiang University\\
 $^3$State Key Laboratory of Blockchain and Security, Zhejiang University \\
 $^4$Hangzhou High-Tech Zone (Binjiang) Institute of Blockchain and Data Security \\[2mm]
 {\tt\small \{qianz,sjie,wanyy,huiqiong\_wang,brooksong\}@zju.edu.cn}
}

\begin{document}
\maketitle
\begin{abstract}
Diffusion models have recently emerged as the dominant approach in visual generation tasks. However, the lengthy denoising chains and the computationally intensive noise estimation networks hinder their applicability in low-latency and resource-limited environments. Previous research has endeavored to address these limitations in a decoupled manner, utilizing either advanced samplers or efficient model quantization techniques. In this study, we uncover that quantization-induced noise disrupts directional estimation at each sampling step, further distorting the precise directional estimations of higher-order samplers when solving the sampling equations through discretized numerical methods, thereby altering the optimal sampling trajectory. To attain dual acceleration with high fidelity, we propose a sampling-aware quantization strategy, wherein a Mixed-Order Trajectory Alignment technique is devised to impose a more stringent constraint on the error bounds at each sampling step, facilitating a more linear probability flow. Extensive experiments on sparse-step fast sampling across multiple datasets demonstrate that our approach preserves the rapid convergence characteristics of high-speed samplers while maintaining superior generation quality. Code is publicly available at: \url{https://github.com/TaylorJocelyn/Sampling-aware-Quantization}.
\end{abstract}    
\section{Introduction}
\label{sec:intro}

Diffusion models \cite{sohl2015deep, ho2020denoising, song2019generative, song2020score} have demonstrated remarkable competitiveness in mainstream generative tasks \cite{wang2023sinsr, lugmayr2022repaint, zhang2023inversion, singer2022make, bilovs2022modeling, lee2022proteinsgm}. By harnessing the power of intricate posterior probability modeling and stable training regimes, these models effectively circumvent mode collapse, attaining superior generation fidelity and diversity when compared to GANs \cite{aggarwal2021generative} and VAEs \cite{kingma2021variational}. However, two primary computational bottlenecks impede the scalability and real-world applicability of diffusion models: the prolonged denoising chains~\cite{ho2020denoising}, and the resource-intensive noise estimation networks. 
\captionsetup[subfigure]{font=small, skip=-5pt, justification=centering}
\captionsetup{justification=centering, singlelinecheck=false} 
\begin{figure}[t]
    \centering
    \hspace*{-7pt} 
    \includegraphics[width=0.492\textwidth]{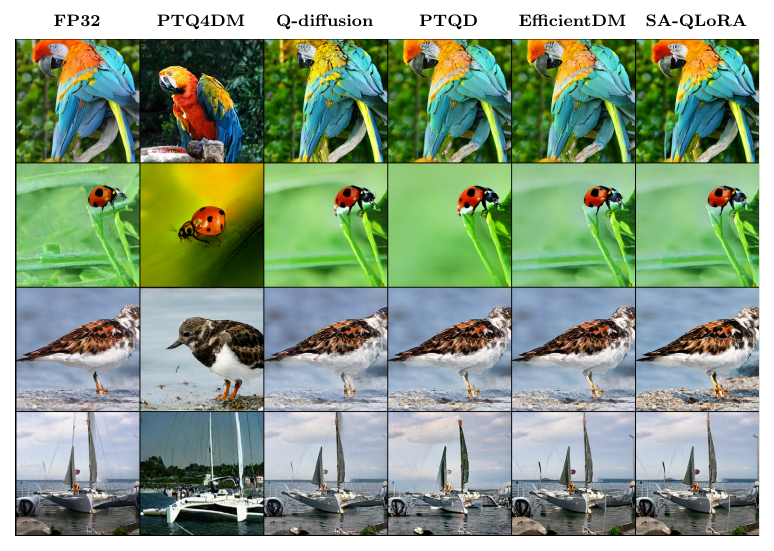}
    \hspace*{\fill} 
    \caption{\justifying Comparison of generated samples on the ImageNet 256×256 dataset between full-precision LDM-4 and its quantized versions using PTQ4DM, Q-diffusion, PTQD, EfficientDM and our proposed SA-QLoRA).}
    \label{figure:1}
   \vspace{-2em} 
\end{figure}
To address the former, advanced samplers \cite{song2020denoising, lu2022dpm, lu2022dpmplus, zhou2024fast} have been developed to achieve efficient sampling trajectories with accurate approximations for stochastic differential equations (SDEs) \cite{dockhorn2021score, liu2022pseudo} and ordinary differential equations (ODEs) \cite{lu2022fast}. Regarding the latter, techniques such as 
quantization \cite{he2023efficientdm, shang2023post, li2023q} have been employed to compress the noise estimation network, reducing both model size and the time and memory costs per iteration. While these two categories of approaches are generally regarded as separate components for accelerating diffusion, quantization errors disrupt the sampler's directional evaluation, degrading high-speed sampling performance. This study, therefore, seeks to develop a sampling-aware quantization strategy aimed at achieving high-fidelity dual acceleration in diffusion models.

Specifically, quantization facilitates efficient model compression and inference acceleration by converting a pre-trained FP32 network into fixed-point networks with lower bit-width representations for weights and activations. This numerical transformation truncates the fractional components, inducing a distribution shift in both weights and activations. Consequently, in quantized diffusion models, quantization noise perturbs each directional estimation step, introducing directional bias. This issue is particularly pronounced in high-order samplers, where, within each interval \((t_{i-1}, t_i)\) of the time schedule \(\{t_i\}_{i=N}^1\), multiple directional estimations are required along the intermediate step sequence \(\{s_j \mid s_j \in (t_{i-1}, t_i), \, j = 1, \dots, n\}\) to establish a collectively estimated direction. Quantization noise perturbs each estimation at intermediate step \(s_j\), substantially distorting the jointly estimated direction (refer to Sec.~\ref{sec:3} for a detailed analysis). The multi-intermediate-step joint directional estimation, intrinsic to high-order samplers, aims to minimize truncation errors arising from the numerical solution of the continuous reverse diffusion equation, thus enabling efficient sparse-step trajectory sampling. However, quantization noise not only hinders the sampler's rapid convergence potential but may also transform the stable probability flow ODE, designed for acceleration, into a variance-exploding SDE, inducing trajectory diffusion.

To mitigate the disruption caused by quantization algorithms on sampling acceleration and to optimally harness the advantages of both acceleration strategies, we propose a sampling-aware quantization technique. This method employs a \textit{Mixed-Order Trajectory Alignment} strategy, thereby fostering a more linear probability flow through the quantization process. In essence, this approach imposes a more stringent constraint on the error bounds at each sampling step (see Sec.~\ref{sec:3_3}), effectively curbing error accumulation and averting sampling diffusion. Subsequently, we integrate the proposed sampling-aware quantization method atop the PTQ and QLoRA baselines, allowing for adaptation to various quantization bit-width requirements. Experiments on rapid sampling with sparse steps across diverse datasets reveal that our approach preserves the swift convergence capabilities of high-order samplers while upholding exceptional generative quality. In conclusion, our main contributions in this work are summarized as follows:
\begin{itemize}
\item We introduce a pioneering Sampling-Aware Quantization framework for diffusion models, examining the influence of quantization errors on rapid sampling through the lens of sampling acceleration principles, and presenting the Mixed-Order Trajectory Alignment strategy to foster a more linear probability flow during quantization.


\item To accommodate diverse quantization bit-width requirements, we tailor sampling-aware quantization to post-training quantization and QLoRA, culminating in the development of SA-PTQ and SA-QLoRA variants.

\item Extensive experiments with sparse-step fast sampling across multiple datasets demonstrate that our method preserves the rapid convergence properties of high-speed samplers while maintaining superior generation quality.
\end{itemize}

\begin{figure*}[h!]
    \centering
    \begin{subfigure}{0.23\textwidth}
        \centering
        \includegraphics[width=\textwidth]{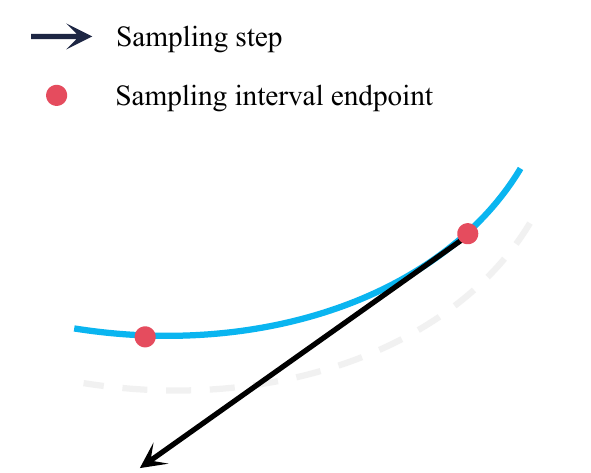}
        \vphantom{g}
        \caption{}
        \label{figure:2_a}
    \end{subfigure}
    \begin{subfigure}{0.23\textwidth}
        \centering
        \includegraphics[width=\textwidth]{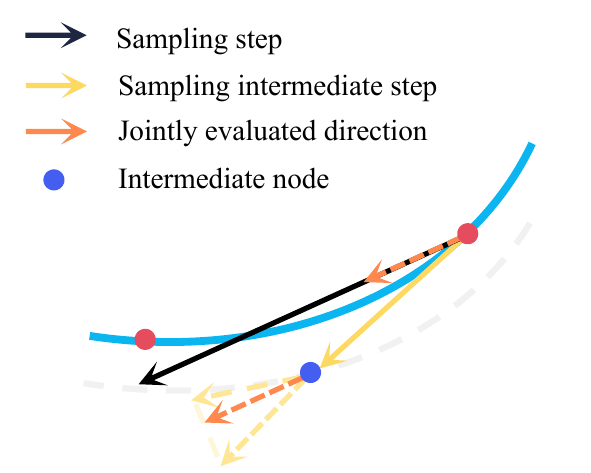}
        \vphantom{g}
        \caption{}
        \label{figure:2_b}
    \end{subfigure}
    \begin{subfigure}{0.255\textwidth}
        \centering
        \includegraphics[width=\textwidth]{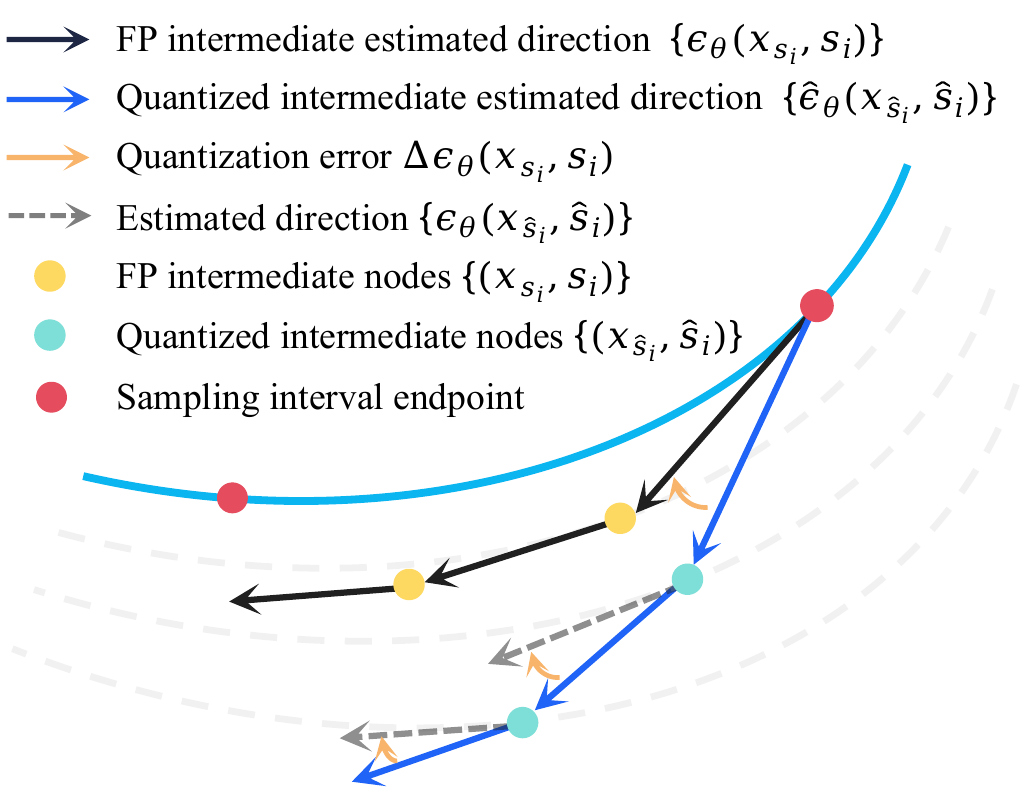}
        \vphantom{g}
        \caption{}
        \label{figure:2_c}
    \end{subfigure}
    \begin{subfigure}{0.23\textwidth}
        \centering
        \includegraphics[width=\textwidth]{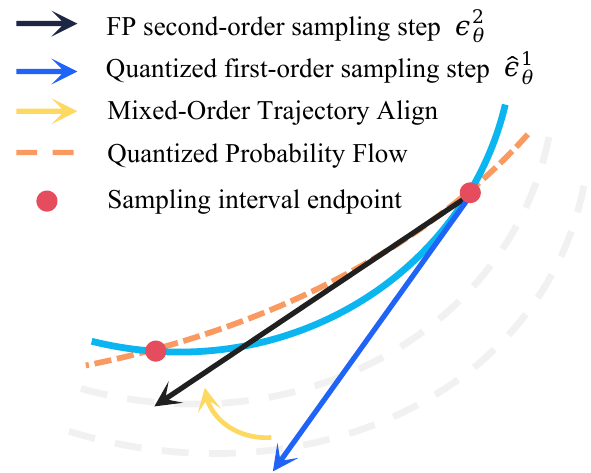}
        \vphantom{g}
        \caption{}
        \label{figure:2_d}
    \end{subfigure}
    \vspace{2pt}
    \caption{\justifying Direction estimation in reverse diffusion sampling. \textbf{(a)} The first-order sampler performs a single direction estimation at the beginning of the sampling interval. \textbf{(b)} The second-order sampler refines the direction estimation by evaluating additional intermediate steps within the interval. \textbf{(c)} Quantization errors lead to deviations in direction estimation, causing the intermediate steps in high-order samplers to drift over time, ultimately impacting the final direction estimation. \textbf{(d)} Our proposed Mixed-Order Trajectory Alignment achieves a more linearized probability flow.}
    \label{figure:2}
\end{figure*}
\section{Related Work}
\subsection{Efficient Diffusion Models}
Diffusion models achieve impressive generation quality and diversity but are constrained by generation speed. Existing research accelerates diffusion from two main perspectives: optimizing generation trajectories for greater efficiency and compressing the noise estimation network to reduce the computational cost per iteration. In the first category, some works \cite{kingma2021variational, kong2021fast} use learning-based methods to optimize $\sigma(t)$ for efficient generation trajectories. By adjusting the signal-to-noise ratio distribution, they minimize the variance of the variational lower bound (VLB), enabling the model to approximate the target distribution more stably and efficiently, reducing unnecessary noise accumulation. Other works focus on learning-free samplers \cite{lu2022dpm, lu2022dpmplus, xu2023restart, zhou2024fast}, achieving high-precision numerical approximations of the sampling SDE and ODE, allowing larger sampling step sizes while controlling discretization truncation errors. Further related work on sampling acceleration can be found in Appendix.~\ref{appendix:B}. The second category applies model lightweighting paradigms such as distillation \cite{salimans2022progressive, zheng2024trajectory}, pruning \cite{fang2024structural, castells2024ld}, and quantization \cite{li2023q, shang2023post, he2023efficientdm}. Distillation and pruning require extensive parameter training, whereas quantization, widely adopted for deployment, can be implemented with minimal or no additional training by adjusting only a few quantization parameters. In this paper, we focus on the joint optimization of learning-free sampling and quantization. 



\subsection{Model Quantization}

Quantization is a mainstream technique for model compression and computational acceleration, converting FP32 weights and activations to low-bit fixed-point counterparts. Two primary approaches exist: post-training quantization (PTQ) \cite{nagel2020up, ding2022towards}, which determines quantization parameters by minimizing the MSE or cross-entropy \cite{nagel2106white} between pre- and post-quantization tensors, and quantization-aware training (QAT) \cite{lin2024awq, liu2023llm}, which trains the network to learn optimal parameters by modeling quantization noise. Common asymmetric quantization involves three parameters: scale factor $s$, zero point $z$, and quantization bit-width $b$. A floating-point value x is quantized to a fixed-point value $x_{int}$ through the preceding parameters:
\begin{equation}
x_{\text{int}} = \text{clamp}\left(\left\lfloor \frac{x}{s} \right\rceil + z, 0, 2^{b}\right),
\label{eq:1}
\end{equation}
where $\lfloor \cdot \rceil$ denotes rounding, with $\text{clamp}$ as truncation.

\subsection{Diffusion Model Quantization}
Current research on diffusion model quantization remains relatively sparse. PTQ4DM \cite{shang2023post} introduces a normal-distribution time-step calibration method, but in their work the experiments are only conducted on low-resolution datasets. Q-Diffusion \cite{li2023q} presents a time-step-aware calibration and shortcut-splitting quantization for U-Net. PTQD \cite{he2024ptqd} applies a PTQ error correction method that requires additional statistical parameters during inference. TDQ \cite{so2024temporal} proposes a time-dynamic quantization strategy with a trained auxiliary network to estimate quantization parameters across time steps. EfficientDM \cite{he2023efficientdm} develops QALoRA for low-bit quantization, though additional training is required. 
These studies primarily focus on adapting traditional quantization techniques to the multi-time-step framework of diffusion models, while overlooking the inevitable impact of quantization noise on high-speed sampling. In contrast, our work integrates sampling acceleration to formulate a high-fidelity dual-acceleration scheme.
\begin{figure*}[h!]
    \centering
    \begin{subfigure}{0.495\textwidth} 
        \centering
        \includegraphics[width=\textwidth]{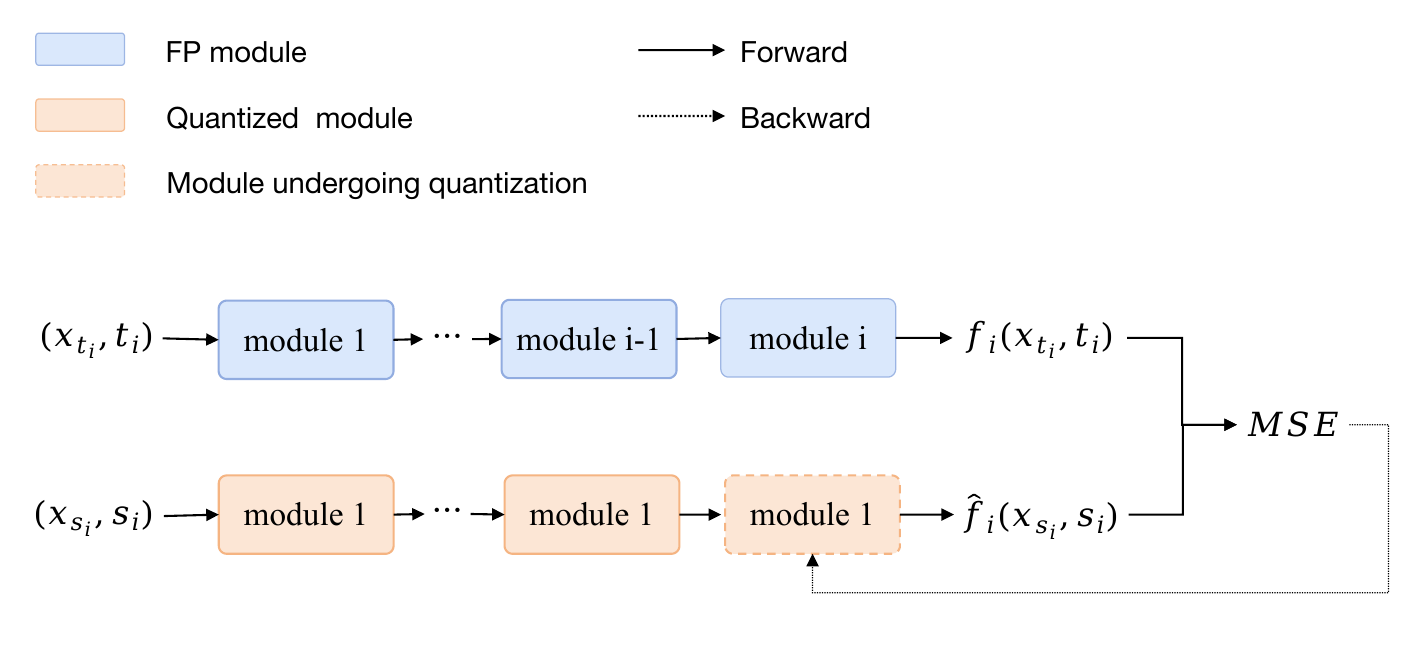}
        \caption{}
        \label{figure:3_a}
    \end{subfigure}
    \begin{subfigure}{0.495\textwidth} 
        \centering
        \includegraphics[width=\textwidth]{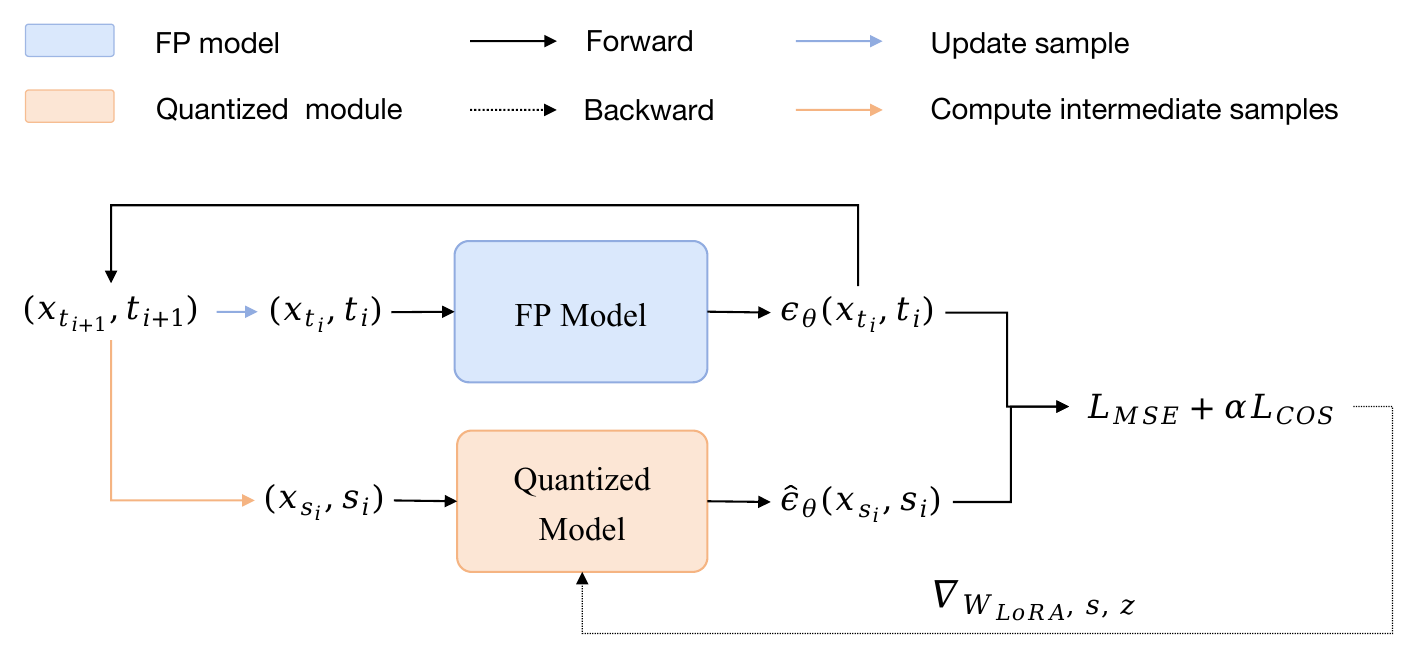}
        \caption{}
        \label{figure:3_b}
    \end{subfigure}
    \vspace{2pt}
    \caption{\justifying Sampling-aware quantization workflow. (a) Module-level reconstruction process employed in SA-PTQ, where $\hat{f}_i(\cdot)$ denotes the module undergoing quantization and reconstruction. (b) Basic fine-tuning workflow in SA-QLoRA, where LoRA weights $W_{LoRA}$ and quantization parameters $s, z$ are iteratively updated after each sampling step.}
    \label{figure:3}
\end{figure*}
\section{Preliminaries} \label{sec:3}
\subsection{Diffusion Models}
Diffusion models progressively adds isotropic Gaussian noise to real data $x_0$ and learn the denoising process by approximating the posterior probability distribution $\{p(x_{t-1}|x_{t})\}_{t=T}^{1}$. The forward process can be modeled as an SDE:
\begin{equation}
\mathrm{d}\mathbf{x} = \mathbf{f}(\mathbf{x}, t) \, \mathrm{d}t + g(t) \, \mathrm{d}\mathbf{w},
\label{eq:2}
\end{equation}
where $\mathbf{w}$ is the standard Wiener process (\textit{a.k.a.}, Brownian motion), $\mathbf{f}(\cdot, t):\mathbb{R}^d \rightarrow \mathbb{R}^d$ is a vector-valued function called the drift coefficient of $\mathbf{x}(t)$, and $g(\cdot):\mathbb{R} \rightarrow \mathbb{R}$ is a scalar function known as the diffusion coefficient of $\mathbf{x}(t)$. 

The predominant sampling methodologies are categorized into deterministic and stochastic sampling. Stochastic sampling follows Anderson’s reverse-time SDE:
\begin{equation}
\mathrm{d}\mathbf{x} = \left[ \mathbf{f}(\mathbf{x}, t) - g(t)^2 \nabla_{\mathbf{x}} \log p_t(\mathbf{x}) \right] \mathrm{d}t + g(t) \mathrm{d}\bar{\mathbf{w}},
\label{eq:3}
\end{equation} 
where $\nabla_{\mathbf{x}} \log p_t(\mathbf{x})$ is the score function \cite{bao2022analytic}, $\bar{\mathbf{w}}$ is a standard Wiener process when time flows backwards from T to 0, and dt is an infinitesimal negative timestep. 
Deterministic sampling follows the \textit{probability flow} ODE:
\begin{equation}
\mathrm{d}\mathbf{x} = \left[ \mathbf{f}(\mathbf{x}, t) - \frac{1}{2} g(t)^2 \nabla_{\mathbf{x}} \log p_t(\mathbf{x}) \right] \mathrm{d}t,
\label{eq:4}
\end{equation}
which shares the same marginal probability as the reverse-time SDE. By eliminating the need for random noise sampling during the generation process, it achieves a more stable and smoother trajectory, facilitating integration with efficient numerical solvers. 

To estimate the score $\nabla_{\mathbf{x}} \log p_t(\mathbf{x})$ in Eqn.~\eqref{eq:3} and Eqn.~\eqref{eq:4}, it is common to train a time-independent score-based model $\mathbf{s}_{\theta}(\mathbf{x}, t)$, which is linearly related to the noise estimation network $\boldsymbol{\epsilon}_{\theta}$:
\begin{equation}
\mathrm{\mathbf{s_\theta}}(\mathbf{x}, t)\triangleq - \frac{\mathrm{\boldsymbol{\epsilon}}_\theta(\mathbf{x}_t, t)}{\sigma_t},
\label{eq:5}
\end{equation} 
where $\sigma_t$ is the standard deviation of $p(\mathbf{x}_t|\mathbf{x}_0)$, referred to as the noise schedule.
\subsection{High-Speed Sampling} \label{sec:3_2}
The inverse sampling process is equivalent to substituting Eqn.~\eqref{eq:5} into Eqn.~\eqref{eq:4} for integration, i.e., given an initial sample $\mathbf{x}_s$ at time $s>0$, the solution $\mathbf{x}_t$ at each $t<s$ satisfies:
\begin{equation}
\mathbf{x}_t = \mathbf{x}_s + \int_{s}^{t} 
\left( f(\mathbf{x}_{\tau}, \tau) + \frac{g(t)^2 \boldsymbol{\epsilon}_{\theta}(\mathbf{x}_{\tau}, \tau)}{2 \sigma_{\tau}} \right) \, \mathrm{d}\tau
\label{eq:6}
\end{equation}
It is evident that $\boldsymbol{\epsilon}_{\theta}(\mathbf{x}_{\tau}, \tau)$ directly affecting the sampling direction. Following the setup of DDPM \cite{ho2020denoising}, where $p(\mathrm{\mathbf{x}}_t | \mathrm{\mathbf{x}}_0) = \mathcal{N}(\alpha_t \mathrm{\mathbf{x}}_0, \sigma_t^2 \bm{I})$, derive the expressions for $\mathbf{f}(\mathrm{\mathbf{x}}, t)$ and g(t). Then, define $\lambda = log(\frac{\alpha_t}{\sigma_t})$, and further change the subscripts of $\mathrm{\mathbf{x}}$ and $\boldsymbol{\epsilon}_{\theta}$ from $t$ to $\lambda$, where $\mathrm{\mathbf{x}}_{\lambda}$ denotes $\mathrm{\mathbf{x}}_{t_{\lambda}(\lambda)}$, resulting in the following integration \cite{lu2022dpm}: 
\begin{equation}
    \mathrm{\mathbf{x}}_t = \frac{\alpha_t}{\alpha_s} \mathrm{\mathbf{x}}_s + \alpha
    _t \int _{\lambda_s} ^{\lambda_t} e^{-\lambda} \boldsymbol{\epsilon}_{\theta}(\mathrm{\mathbf{x}}_{\lambda}, \lambda) \, \mathrm{d}\lambda
    \label{eq:7}
\end{equation}
Due to the intractability of nonlinear integration in nonlinear networks $\boldsymbol{\epsilon}_{\theta}(\mathbf{x}_{\lambda}, \lambda)$, numerical approximation of the sampling direction in continuous equations is achieved by performing a high-order expansion of $\boldsymbol{\epsilon}_{\theta}(\mathbf{x}_{\lambda}, \lambda)$ at $\lambda_s$ \cite{lu2022dpm}:
\begin{footnotesize}
\begin{align}
\mathbf{x}_t & = \frac{\alpha_t}{\alpha_s} \mathbf{x}_s -\alpha_t \sum_{n=0}^{k-1} \boldsymbol{\epsilon}_{\theta}^{(n)}(\mathbf{x}_{\lambda_s}, \lambda_s) \int_{\lambda_s}^{\lambda_t} 
e^{-\lambda} \cdot \frac{(\lambda - \lambda_s)^n}{n!}\, \mathrm{d}\lambda \notag \\
& + \mathcal{O}((\lambda_t-\lambda_s)^{k+1}),
\label{eq:8}
\end{align}
\end{footnotesize}
\normalsize
\noindent where $\boldsymbol{\epsilon}_{\theta}^{(n)}(\mathbf{x}_{\lambda_s}, \lambda_s) = \frac{\mathrm{d}^n\boldsymbol{\epsilon}_{\theta}^{(n)}(\mathbf{x}_{\lambda}, \lambda)}{\mathrm{d}{\lambda}^n} $ is the $n$-th order total derivative of w.r.t. $\lambda$. In practice, the $k$-th order expansion involves selecting $k-1$ intermediate points $\{\lambda_i\}_{i=0}^{k-2}$ within $(\lambda_s, \lambda_t)$, and using the corresponding $\boldsymbol{\epsilon}_{\theta}(\mathbf{x}_{\lambda_i}, \lambda_i)$ to achieve a more accurate approximation of the sampling direction (more details in Appendix A.1), as shown in Fig.~\ref{figure:2_b}. Under the condition of limited truncation error, acceleration is achieved by increasing the sampling step size.

\subsection{Pre-analysis: Quantization Error Interference in High-Speed Sampling} \label{sec:3_3}
The quantized network $\boldsymbol{\hat{\epsilon}}_{\theta}$ inevitably introduces quantization noise $\Delta\boldsymbol{\epsilon}_{\theta}$, resulting in a deviation in the numerical integration in Eqn.~\eqref{eq:8} as follows:
\begin{footnotesize}
\begin{align}
\mathbf{x}_t & = \frac{\alpha_t}{\alpha_s} \mathbf{x}_s -\alpha_t \sum_{n=0}^{k-1} \boldsymbol{\hat{\epsilon}}_{\theta}^{(n)}(\mathbf{x}_{\lambda_s}, \lambda_s) \int_{\lambda_s}^{\lambda_t} 
e^{-\lambda} \cdot \frac{(\lambda - \lambda_s)^n}{n!}\, \mathrm{d}\lambda \notag \\
& + \mathcal{O}((\lambda_t-\lambda_s)^{k+1}) \notag \\
& = \frac{\alpha_t}{\alpha_s} \mathbf{x}_s - \alpha_t \sum_{n=0}^{k-1} \boldsymbol{\epsilon}_{\theta}^{(n)}(\mathbf{x}_{\lambda_s}, \lambda_s) \int_{\lambda_s}^{\lambda_t} 
e^{-\lambda} \cdot \frac{(\lambda - \lambda_s)^n}{n!}\, \mathrm{d}\lambda \notag \\
& + \sum_{n=0}^{k-1} \Delta\boldsymbol{\epsilon}_{\theta}^{(n)}(\mathbf{x}_{\lambda_s}, \lambda_s) \int_{\lambda_s}^{\lambda_t} 
e^{-\lambda} \cdot \frac{(\lambda - \lambda_s)^n}{n!}\, \mathrm{d}\lambda \notag \\
& + \mathcal{O}((\lambda_t-\lambda_s)^{k+1}) \,
\label{eq:9}
\end{align}
\end{footnotesize}
\normalsize
where the deviation of the derivative sequence $\{ \Delta{\boldsymbol{\epsilon}}_{\theta}^{(n)}(\mathbf{x}_{\lambda_s}, \lambda_s) \}_{n=0}^{k-1}$ essentially implies a positional shift of the sampling intermediate points (as described in Sec.~\ref{sec:3_2}), along with a directional estimation shift at these intermediate points, as shown in Fig.~\ref{figure:2_c}. 
\normalsize
We denote $\Delta_{quant} = \sum_{n=0}^{k-1} \Delta\boldsymbol{\epsilon}_{\theta}^{(n)}(\mathbf{x}_{\lambda_s}, \lambda_s) \int_{\lambda_s}^{\lambda_t} 
e^{-\lambda} \cdot \frac{(\lambda - \lambda_s)^n}{n!}\, \mathrm{d}\lambda$ as the quantization cumulative error term, and $\Delta_{disc}$ as the discretization truncation error term, with $\Delta{\boldsymbol{\epsilon}}_{\theta}^{(n)}(\mathbf{x}_{\lambda_s}, \lambda_s)$ simplified to $\delta$. As analyzed in Appendix A.2, the total error upper bound for the numerical integration in Eqn.~\eqref{eq:9}, utilizing the $(k-1)$-th order expansion term, can be expressed as:
\begin{align}
    \mathcal{L}_{\Delta} & = \mathcal{L}_{{\Delta}_{quant}} + \mathcal{L}_{{\Delta}_{disc}} \notag \\
    & = \mathcal{O}(\delta \cdot e^{-\lambda_s} \cdot (\lambda_t - \lambda_s)) + \mathcal{O}((\lambda_t - \lambda_s)^{k+1})
\end{align}

It is evident that the quantization cumulative error, controlled directly by $\delta$, dominates the total error, significantly impacting the rapid convergence of ODE sampling. This cumulative effect is exacerbated by the model’s nonlinearity, which amplifies the propagation of quantization errors through higher-order terms. To address this issue, we redesign the quantization scheme to learn a more linear probability flow, bringing $\delta$ closer to the order of $\mathcal{O}(\lambda_t - \lambda_s)$ to further constrain the error bound and promote convergence.

\section{Methodology}
In this section, we introduce an innovative Sampling-Aware Quantization framework. We begin by presenting the core component—the Mixed-Order Trajectory Alignment strategy—in Sec.~\ref{sec:4_1}, using the DPM-Solver sampler as a case study. To facilitate 8-bit quantization, we propose a sampling-aware post-training quantization method in Sec.~\ref{sec:4_2}. Furthermore, to accommodate lower-bit quantization requirements (e.g., W4A4), this approach is extended in Sec.~\ref{sec:4_3} with a sampling-aware Quantized Low-Rank Adaptation (QLoRA) method. Finally, the adaptation scheme for more generalized samplers is presented in Sec.~\ref{sec:4_4}, boosting the framework's versatility.

\subsection{Mixed-Order Trajectory Alignment} \label{sec:4_1}

High-fidelity quantization for diffusion models is typically achieved by aligning the sampling trajectory of the full-precision model with that of the quantized counterpart.

\noindent\textbf{Sampling Trajectory.} In the reverse diffusion process, the intermediate sample set \(\{\mathrm{\mathbf{x}}_t\}_{t=T}^0\), obtained by numerically integrating the sampling equation from an initial point \(\mathrm{\mathbf{x}}_T \sim \mathcal{N}(\mathbf{0}, \sigma^2 \bm{I})\) along a time schedule \(\{t_i\}_{i=T}^0\), is conventionally referred to as the sampling trajectory. Thus, the sampling trajectory is governed by three primary factors: the sampler, which defines the numerical solution method; the initial sampling point \(\mathrm{\mathbf{x}}_T\); and the time schedule \(\{t_i\}_{i=T}^0\), which can be mapped to a noise schedule \(\{\sigma_i\}_{i=T}^0\). As detailed in Sec.~\ref{sec:3_2}, \(\boldsymbol{\epsilon}_{\theta}(\mathrm{\mathbf{x}}_t, t)\) dictates the sampling direction at each step \(t\). Consequently, once the aforementioned conditions are specified, the sampling trajectory becomes fully determined, with each sampling direction at each step uniquely defined. This establishes a one-to-one correspondence between the direction sequence \(\{\boldsymbol{\epsilon}_{\theta}(\mathrm{\mathbf{x}}_t, t)\}_{t=T}^0\) and the sample trajectory sequence \(\{\mathrm{\mathbf{x}}_t\}_{t=T}^0\). Therefore, we clarify that the trajectory aligned in this work is, in essence, the direction sequence \(\{\boldsymbol{\epsilon}_{\theta}(\mathrm{\mathbf{x}}_t, t)\}_{t=T}^0\). Accordingly, the core quantization objective can be formulated as:

\[
\hat{\epsilon}_{\theta} = \mathcal{Q}(\epsilon_{\theta}, s, z)
\label{eq:11}
\]

\[
\arg \min_{s, z} \mathbb{E}_{(\mathrm{\mathbf{x}}_t, t) \sim \mathcal{D}} \| \boldsymbol{\epsilon}_{\theta}(\mathrm{\mathbf{x}}_t, t) - \hat{\boldsymbol{\epsilon}}_{\theta}(\mathrm{\mathbf{x}}_t, t) \|^2
\label{eq:12}
\]
Here, \(\mathcal{Q}\) represents the quantization function, \(s\) and \(z\) denote the quantization parameters, scale and zero-point respectively, and \(\mathcal{D}\) refers to the sampling distribution of the calibration dataset.

As analyzed in Sec.~\ref{sec:3_2}, the quantization noise \(\Delta\boldsymbol{\epsilon}_{\theta}\) impinges upon the directional estimation of high-speed samplers, particularly exerting cumulative effects on higher-order samplers, which necessitate multiple evaluations of higher-order derivatives, ultimately resulting in trajectory deviation.
To counteract the swift accumulation of errors, we employ mixed-order trajectory quantization to foster a more linear probability flow. As detailed in Appendix A.1, for sampling within the interval \((\lambda_{t_i-1}, \lambda_{t_i})\), a first-order sampler directly evaluates \(\boldsymbol{\epsilon}_{\theta}(\mathrm{\mathbf{x}}_{\lambda_{t_{i-1}}}, \lambda_{t_{i-1}})\) to determine the sampling direction. In contrast, a \(k\)-order sampler generates \(k - 1\) intermediate points \(\{s_i\}_{i=0}^{k-2}\) and evaluates \(\boldsymbol{\epsilon}_{\theta}(\mathrm{\mathbf{x}}_{\lambda_{s_{i}}}, \lambda_{s_{i}})\) at each point \(s_i\), iteratively refining the direction based on \(\boldsymbol{\epsilon}_{\theta}(\mathrm{\mathbf{x}}_{\lambda_{t_{i-1}}}, \lambda_{t_{i-1}})\). Each \(\boldsymbol{\epsilon}_{\theta}(\mathrm{\mathbf{x}}_{\lambda_{s_{i}}}, \lambda_{s_{i}})\) encodes higher-order derivative information, thereby enhancing the directional precision of the sampling process. Drawing inspiration from this, we achieve mixed-order trajectory quantization by aligning the quantized first-order sampling direction trajectory \(\{\boldsymbol{\hat{\epsilon}}_{\theta}(\mathrm{\mathbf{x}}_{\lambda_{t}}, \lambda_{t})\}_{t=T}^{0}\) with the full-precision higher-order sampling direction trajectory \(\{\boldsymbol{\epsilon}_{\theta}(\mathrm{\mathbf{x}}_{\lambda_{s}}, \lambda_{s})\}_{s \in \mathcal{S}}\) at the intermediate nodes.
The quantization objective is defined as:
\begin{equation}
\displaystyle
\hat{\epsilon}_{\theta} = \mathcal{Q}(\epsilon_{\theta}, s, z)
\label{eq:13}
\end{equation}
\begin{equation}
\displaystyle
\arg \min_{s, z} \mathbb{E}_{(\mathrm{\mathbf{x}}_t, t) \sim \mathcal{D}, (\mathrm{\mathbf{x}}_s, s) \sim \mathcal{S}} \| \boldsymbol{\hat{\epsilon}}_{\theta}(\mathrm{\mathbf{x}}_{\lambda_s}, \lambda_s) - \boldsymbol{\epsilon}_{\theta}(\mathrm{\mathbf{x}}_{\lambda_t}, \lambda_t) \|^2
\label{eq:14}
\end{equation}

For example, taking DPM-Solver sampling as a case study. Given an initial value $\mathrm{\mathbf{x}}_T$ and $M + 1$ time steps $\{t_i\}_{i=0}^M$ decreasing from $t_0 = T$ to $t_M = 0$. Starting with $\tilde{\mathrm{\mathbf{x}}}_{t_0} = x_T$, the DPM-Solver-1 sampling sequence $\{\tilde{\mathrm{\mathbf{x}}}_{t_i}\}_{i=1}^M$ is computed iteratively as follows:
\begin{equation}
\tilde{\mathrm{\mathbf{x}}}_{t_i} = \frac{\alpha_{t_i}}{\alpha_{t_{i-1}}} \tilde{\mathrm{\mathbf{x}}}_{t_{i-1}} - \sigma_{t_i} \left( e^{h_i} - 1 \right) \boldsymbol{\epsilon}_\theta(\tilde{\mathrm{\mathbf{x}}}_{t_{i-1}}, t_{i-1}),
\end{equation}
where $h_i = \lambda_{t_i} - \lambda_{t_{i-1}}. $ The corresponding sampling formula for DPM-Solver-2 is given in Alg.~\ref{alg:dpm-solver-2}. 

\begin{algorithm}[H]
\caption{DPM-Solver-2} \label{alg:dpm-solver-2}
\begin{algorithmic}[1]
\Require Initial value $\mathrm{\mathbf{x}}_T$, time steps $\{t_i\}_{i=0}^M$, model $\epsilon_\theta$
\State $\tilde{\mathrm{\mathbf{x}}}_{t_0} \gets \mathrm{\mathbf{x}}_T$
\For{$i \gets 1$ to $M$}
    \State $s_i \gets t_\lambda \left( \frac{\lambda_{t_{i-1}} + \lambda_{t_i}}{2} \right)$
    \State $\mathrm{\mathbf{u}}_i \gets \frac{\alpha_{s_i}}{\alpha_{t_{i-1}}} \tilde{\mathrm{\mathbf{x}}}_{t_{i-1}} - \sigma_{s_i} \left( e^{\frac{h_i}{2}} - 1 \right) \boldsymbol{\epsilon}_\theta(\tilde{\mathrm{\mathbf{x}}}_{t_{i-1}}, t_{i-1})$
    \State $\tilde{\mathrm{\mathbf{x}}}_{t_i} \gets \frac{\alpha_{t_i}}{\alpha_{t_{i-1}}} \tilde{\mathrm{\mathbf{x}}}_{t_{i-1}} - \sigma_{t_i} \left( e^{h_i} - 1 \right) \boldsymbol{\epsilon}_\theta(\mathrm{\mathbf{u}}_i, s_i)$
\EndFor
\State \Return $\tilde{\mathrm{\mathbf{x}}}_{t_M}$
\end{algorithmic}
\end{algorithm}

By aligning the quantized directional term $\boldsymbol{\hat{\epsilon}}_\theta(\tilde{\mathrm{\mathbf{x}}}_{t_{i-1}}, t_{i-1})$ with the full-precision directional term $\boldsymbol{\epsilon}_\theta(\mathrm{\mathbf{u}}_i, s_i)$, we can effectively linearize higher-order sampling trajectories, as illustrated in Fig.~\ref{figure:2_d}.





\subsection{Sampling-Aware Post-Training Quantization} \label{sec:4_2}
For 8-bit quantization, we propose a Sampling-Aware Post-Training Quantization scheme (SA-PTQ), with the process illustrated in Fig.~\ref{figure:3_a}. In alignment with prior work \cite{shang2023post, li2023q, he2024ptqd}, we adopt the widely utilized BRECQ \cite{li2021brecq} as the baseline quantization algorithm, utilizing Adaround for weight quantization while training only the minimal parameter $\alpha$. Building upon the method outlined in Sec.~\ref{sec:4_1}, we develop a novel dual-order trajectory calibration strategy to guide module reconstruction. This strategy comprises two key components: Dual-order Trajectory Sampling and Mixed-Order Trajectory Alignment Calibration.

\noindent\textbf{Dual-Order Trajectory Sampling.} Given a predefined seed $seed$ and time schedule $\{t_i\}_{i=N}^0$, we initiate sampling from an initial sample $\mathrm{\mathbf{x}}_T \sim \mathcal{N}(0, \bm{I})$. Utilizing a first-order sampler, we collect input samples $\{t_i\}_{i=N}^0$ at each noise estimation iteration to establish the first-order trajectory, where $cond$ encapsulates the conditional information. Subsequently, we input the same $\mathrm{\mathbf{x}}_T$ into a second-order sampler, obtaining input samples $\{(\mathrm{\mathbf{x}}_{s_i}, s_i, cond)\}_{s_i \in (t_{i-1}, t_i)}$ at each intermediate point $s_i$ within the interval $(t_{i-1}, t_i)$, thus forming the second-order trajectory.


\noindent\textbf{Mixed-Order Trajectory Alignment Calibration.}
Let $f_i(\cdot)$ represent the $i$-th module within the noise estimation network requiring reconstruction, and $\hat{f}_i(\cdot)$ denote its quantized counterpart. The SA-PTQ approach attains high-fidelity quantization by reconstructing each module independently. The objective for reconstructing each module is formulated as:

\begin{equation}
\hat{f}_i = Adaround(f_i, \alpha)
\label{eq:16}
\end{equation}
\begin{equation}
\arg \min_{\alpha} \mathbb{E}_{(t_j, s_j)} \| f_i(\mathrm{\mathbf{x}}_{t_j}, t_j, cond) - \hat{f}_i(\mathrm{\mathbf{x}}_{s_j}, s_j, cond) \|^2
\label{eq:17}
\end{equation}
  
\subsection{Sampling-Quantization Dual-Aware LoRA} \label{sec:4_3}
To meet the demands of low-bit quantization, we integrate Mixed-Order Trajectory Alignment with QLoRA, introducing the Sampling-Quantization Dual-Aware LoRA (SA-QLoRA) framework. The workflow is depicted in Fig.~\ref{figure:3_b}.

For the basic QLoRA, training is supervised by aligning $\boldsymbol{\hat{\epsilon}}_{\theta}(\mathrm{\mathbf{x}}_{t_i}, t_i)$ and $\boldsymbol{\epsilon}_{\theta}(\mathrm{\mathbf{x}}_{t_i}, t_i)$ at each time step $t_i$ , guiding the optimization of both the LoRA weights $w$ and the quantization parameters $s$ and $z$. The quantization objective can be formulated as:
\begin{equation}
\boldsymbol{\hat{\epsilon}}_{\theta} = \mathit{QLoRA}(\boldsymbol{\epsilon}_{\theta}, w, s, z)
\label{eq:18}
\end{equation}
\begin{equation}
\arg \min _{w, s, z} \mathbb{E}_{(\mathrm{\mathbf{x}}_{t_i}, t_i) \sim \mathcal{D}} \| \boldsymbol{\hat{\epsilon}}_{\theta}(\mathrm{\mathbf{x}}_{t_i}, t_i) - \boldsymbol{\epsilon}_{\theta}(\mathrm{\mathbf{x}}_{t_i}, t_i) \|^2
\label{eq:19}
\end{equation}

In light of the analysis in Sec.~\ref{sec:3_2}, $\boldsymbol{\epsilon}_{\theta}(\mathrm{\mathbf{x}}_{t_i}, t_i)$ directly determines the sampling direction, we introduce an additional directional constraint $\mathcal{L}_{COS}$ to enhance the Mixed-Order Trajectory Alignment constraint $\mathcal{L}_{MOTA}$ in supervising QLoRA training, with the resulting training objective formulated as:
\begin{equation}
\mathcal{L}_{COS} = 1 - \frac{\langle \boldsymbol{\epsilon}_{\theta}(\mathrm{\mathbf{x}}_{t_i}, t_i), \boldsymbol{\hat{\epsilon}}_{\theta}(\mathrm{\mathbf{x}}_{s_i}, s_i) \rangle}{\|\boldsymbol{\epsilon}_{\theta}(\mathrm{\mathbf{x}}_{t_i}, t_i)\| \|\boldsymbol{\hat{\epsilon}}_{\theta}(\mathrm{\mathbf{x}}_{t_i}, t_i)\|}
\label{eq:20}
\end{equation}
\begin{equation}
\mathcal{L}_{MOTA} =  \mathbb{E}_{(t_i, s_i)} \| \boldsymbol{\hat{\epsilon}}_{\theta}(\mathrm{\mathbf{x}}_{s_i}, s_i) - \boldsymbol{\epsilon}_{\theta}(\mathrm{\mathbf{x}}_{t_i}, t_i) \|^2
\label{eq:21}
\end{equation}
\begin{equation}
\arg \min _{w, s, z} \mathcal{L}_{COS} + \mathcal{L}_{MOTA} \, ,
\label{eq:22}
\end{equation}
where the set $\{t_i\}_{i=N}^0$ represents the collection of first-order sampling points, while $\{s_i\}_{i=N}^0$ denotes the additional intermediate evaluation points for second-order sampling.

\subsection{Adaptation to Generalized Samplers} \label{sec:4_4}

As the classical numerical method-based generalized solver for diffusion ODEs, DPM-Solver's \cite{lu2022dpm} sampling-aware quantization adaptation scheme is discussed in Sec.~\ref{sec:4_1}. Additionally, DDIM \cite{song2020denoising} is proven to exhibit identical updates to DPM-Solver-1 \cite{lu2022dpm}, and can thus be regarded as a specific instance of DPM-Solver-1 under a particular noise schedule for sampling lower-order trajectories.

\noindent \textbf{PLMS sampler.} PNDM \cite{liu2022pseudo} decomposes the numerical sampling equation into a gradient part and a transfer part, and defines the pseudo-numerical sampling equation by introducing nonlinear transfer parts $\phi(\cdot)$, as follows:
\begin{align*}
    &\phi(\mathrm{\mathbf{x}}_t, \bm{\epsilon}_{\theta}^{(t)}, t, t - \delta) =
    \frac{\sqrt{\bar{\alpha}_{t - \delta}}}{\sqrt{\bar{\alpha}_t}} \mathrm{\mathbf{x}}_t \\
    &- \frac{(\bar{\alpha}_{t - \delta} - \bar{\alpha}_t)}
    {\sqrt{\bar{\alpha}_t} \left( \sqrt{(1 - \bar{\alpha}_{t - \delta}) \bar{\alpha}_t} + \sqrt{(1 - \bar{\alpha}_t) \bar{\alpha}_{t - \delta}} \right)}
    \bm{\epsilon}_{\theta}^{(t)},
    \tag{21} \label{eq:21}
\end{align*}
where $\bar{\alpha}_{i}$ is a parameter related to the noise schedule, and $\bm{\epsilon}_{\theta}^{(t)}$ is the gradient part that determines the sampling direction. Various well-established numerical methods can be employed to estimate $\bm{\epsilon}_{\theta}^{(t)}$ (e.g., the linear multi-step method applied in \textbf{P}seudo-\textbf{L}inear \textbf{M}ulti-\textbf{S}tep samplers), with different orders of numerical methods corresponding to trajectories of different orders. Therefore, we achieve Mixed-Order Trajectory Alignment by aligning $\bm{\epsilon}_{\theta}^{(t)}$ derived from numerical methods of varying orders.

\section{Experiments}
\subsection{Settings}
\textbf{Benchmarks and Metrics.} We evaluated the proposed SA-PTQ and SA-QLoRA across multiple benchmarks: using LDM \cite{rombach2022high} for class-conditional image generation on ImageNet 256$\times$256; LDM for unconditional image generation on LSUN-Churches 256$\times$256 and LSUN-Bedroom 256$\times$256 \cite{yu2015lsun}; and SD-v1.4 for text-guided image generation on MS-COCO 512$\times$512 \cite{lin2014microsoft}. For the first three benchmarks, we employ metrics such as Fréchet Inception Distance (FID), Sliding Fréchet Inception Distance (sFID), Inception Score (IS), precision, and recall to comprehensively evaluate algorithm performance. For each evaluation, we generate 50k samples and calculate these metrics using the OpenAI's evaluator \cite{dhariwal2021diffusion}, with BOPs (Bit Operations) as the efficiency metric. For the text-to-image benchmark, we further incorporate CLIP-Score to evaluate text-image consistency, generating 30k samples per evaluation round.

\begin{table*}[ht]
\centering
\hfill \caption{\RaggedRight Performance evaluation of class-conditioned image generation on the ImageNet 256 $\times$ 256 dataset using LDM-4 with 20 sampling steps of DPM-Solver-2.}
\small
\resizebox{\textwidth}{!}{%
\begin{tabular}{cccccccccc}
\toprule
\textbf{Model} & \textbf{Method} & \textbf{Bits (W/A)} & \textbf{Size (MB)} & \textbf{BOPs (T)} & \textbf{IS $\uparrow$} & \textbf{FID $\downarrow$} & \textbf{sFID $\downarrow$} & \textbf{Precision $\uparrow$} & \textbf{Recall $\uparrow$} \\ 
\midrule
\multirow{14}{*}{\begin{tabular}[c]{@{}c@{}}LDM-4\\
($scale = 1.5$, \\
$steps = 20$)\end{tabular}}
\multirow{3}{*}{} & FP & 32/32 & 1742.72 & 102.20 & 174.33 & 9.45 & 8.08 & 77.22$\%$ & 52.25$\%$ \\ 
\cmidrule(lr){2-3} \cmidrule(lr){4-5} \cmidrule(lr){6-10}
& PTQ4DM & 8/8 & 436.79 & 8.76 & 115.06 & 11.43 & 12.19 & 60.21$\%$ & 51.26$\%$ \\ 
& Q-diffusion & 8/8 & 436.79 & 8.76 & 120.14 & 10.98 & 11.34 & 63.97$\%$ & 54.34$\%$ \\ 
& PTQD & 8/8 & 436.79 & 8.76 & \textbf{122.46} & 10.76 & 10.58 & 62.08$\%$ & 56.16$\%$\\ 
& SA-PTQ (ours) & 8/8 & 436.79 & 8.76 & 120.71 & \textbf{10.16} & \textbf{9.89} & \textbf{65.39}$\%$ & \textbf{56.97}$\%$ \\ 
\cmidrule(lr){2-3} \cmidrule(lr){4-5} \cmidrule(lr){6-10}
& PTQ4DM & 4/8 & 219.12 & 4.38 & 122.75 & 10.14 & 12.73 & 69.86$\%$ & 51.03$\%$ \\ 
& Q-diffusion & 4/8 & 219.12 & 4.38 & 130.69 & 9.76 & 10.92 & 68.45$\%$ & 51.97$\%$ \\ 
& PTQD & 4/8 & 219.12 & 4.38 & 127.41 & 9.16 & 9.72 & 72.80$\%$ & 50.41$\%$ \\ 
& EfficientDM & 4/8 & 219.12 & 4.38 & 132.70 & 9.91 & 8.76 & 71.05$\%$ & 53.62$\%$ \\  
& SA-QLoRA (ours) & 4/8 & 219.12 & 4.38 & \textbf{140.56} & \textbf{8.55} & \textbf{8.51} & \textbf{73.20$\%$} & \textbf{54.49$\%$} \\  
\cmidrule(lr){2-3} \cmidrule(lr){4-5} \cmidrule(lr){6-10}
& PTQ4DM & 4/4 & 219.12 & 2.19 & - & - & - & - & - \\ 
& Q-diffusion & 4/4 & 219.12 & 2.19 & - & - & - & - & -\\ 
& PTQD & 4/4 & 219.12 & 2.19 & - & - & - & - & - \\ 
& EfficientDM & 4/4 & 219.12 & 2.19 & 225.20 & 17.28 & 13.78 & \textbf{60.33$\%$} & 52.82$\%$ \\  
& SA-QLoRA (ours) & 4/4 & 219.12 & 2.19 & \textbf{242.03} & \textbf{13.73} & \textbf{12.45} & 58.90$\%$ & \textbf{55.38$\%$} \\  
\bottomrule
\end{tabular}%
}
\label{table:1}
\end{table*}

\noindent\textbf{Model and Sampling settings.} For both class-conditional and unconditional image generation, we adopt DPM-Solver-1 and DPM-Solver-2 as the low-order and high-order samplers, respectively, in the SA-PTQ and SA-QLoRA frameworks to achieve mixed-order trajectory alignment. Our primary focus is on two parameters of the generative sampler within LDM: the classifier-free guidance scale $scale$ and the number of sampling steps $steps$. For class-conditional generation, we set {steps=20, scale=1.5}, while for unconditional generation, we set {steps=50}. For text-to-image tasks, we align the native PLMS sampling trajectory with its one-order-reduced counterpart, setting {steps=50, scale=7.5}.

\noindent\textbf{Quantization Settings.} We denote quantization of weights to $x$-bits and activations to $y$-bits as $WxAy$. For further details on the quantization settings and SA-QLoRA fine-tuning, please refer to Appendix.


\subsection{Main Results}
\subsubsection{Class-conditional Generation} 
We first compare our proposed SA-PTQ and SA-QLoRA with previous approaches on class-conditioned image generation task. Specifically, we conduct evaluations on ImageNet 256 $\times$ 256 using a pre-trained LDM-4 model with DPM-Solver-2 over 20 sampling steps. The results are presented in Table 1. In terms of efficiency, configurations $W8A8$, $W4A8$, and $W4A4$ achieve bit compression rates of 3.99x, 7.95x, and 7.95x, respectively, along with bit-operation acceleration rates of 11.47x, 23.33x, and 46.67x. In terms of generation quality, under W8A8 and $W4A8$ configurations, our proposed SA-PTQ and SA-QLoRA consistently demonstrate superior performance across all metrics, achieving the lowest FID and sFID scores of 8.55 and 8.51, respectively. Notably, the FID score is even 0.9 lower than that of the full-precision model. Under the $W4A4$ configuration, previous work \cite{li2023q, shang2023post, he2024ptqd} introduce excessive quantization noise, transforming the originally deterministic probability flow ODE into a variance-exploding SDE, ultimately resulting in generation failure. In contrast, our SA-LoRA demonstrates excellent convergence, with an sFID only 4.37 points higher than the full-precision model.

The consistently strong metrics across various quantization settings confirm that our mixed-order trajectory alignment strategy has, to a certain extent, achieved a more linear probability flow through quantization. Consequently, this approach effectively mitigates the rapid accumulation of high-order sampler errors induced by quantization, thereby preserving outstanding generative performance under fast sampling with sparse trajectories.

\subsubsection{Unconditional Generation}

We then thoroughly evaluate SA-PTQ and SA-QLoRA on unconditional generation tasks, employing the LDM-4 and LDM-8 models across the LSUN-Bedroom and LSUN-Church datasets, respectively. Tab.~\ref{table:2} and Tab.5(in Appendix) indicate that our approach narrows the gap with the full-precision model. 
\begin{table}[h]
\centering
\caption{\justifying Performance comparisons of unconditional image generation on LSUN-Bedroom 256 $\times$ 256.}
\footnotesize
\begin{tabular}{cccccc}
\toprule
\multicolumn{6}{c}{LDM-4 (steps = 50)} \\
\midrule
\textbf{Method} & \textbf{W/A} & \textbf{FID $\downarrow$} & \textbf{sFID $\downarrow$} & \textbf{Prec. $\uparrow$} & \textbf{Rec. $\uparrow$}  \\ 
\midrule
FP & 32/32 & 6.17 & 13.92 & 64.30$\%$ & 51.46$\%$ \\ 
\midrule
PTQD & 8/8 & 10.31 & 15.89 & 56.57$\%$ & 54.15$\%$ \\ 
SA-PTQ & 8/8 & \textbf{9.83} & \textbf{13.85} & \textbf{56.86}$\%$ & \textbf{54.54}$\%$  \\ 
\cmidrule(lr){1-2} \cmidrule(lr){3-6} 
PTQD & 4/8 & 10.96 & 15.42 & 47.83$\%$ & 52.80$\%$ \\ 
EfficientDM & 4/8 & 9.32 & 14.48 & 48.62$\%$ & \textbf{53.02$\%$} \\ 
SA-QLoRA & 4/8 & \textbf{8.79} & \textbf{12.41} & \textbf{50.16}$\%$ & 52.31$\%$  \\ 
\cmidrule(lr){1-2} \cmidrule(lr){3-6} 
PTQD & 4/4 & - & - & - & - \\ 
EfficientDM & 4/4 & 19.30 & 22.63 & \textbf{43.77$\%$} & 40.09$\%$ \\ 
SA-QLoRA & 4/4 & \textbf{14.72} & \textbf{18.19} & 41.84$\%$ & \textbf{45.60}$\%$  \\ 
\bottomrule
\end{tabular}%
\label{table:2}
\end{table}
\begin{figure*}[ht!]
    \centering
    \begin{subfigure}{0.3\textwidth}
        \centering
        \includegraphics[width=\textwidth]{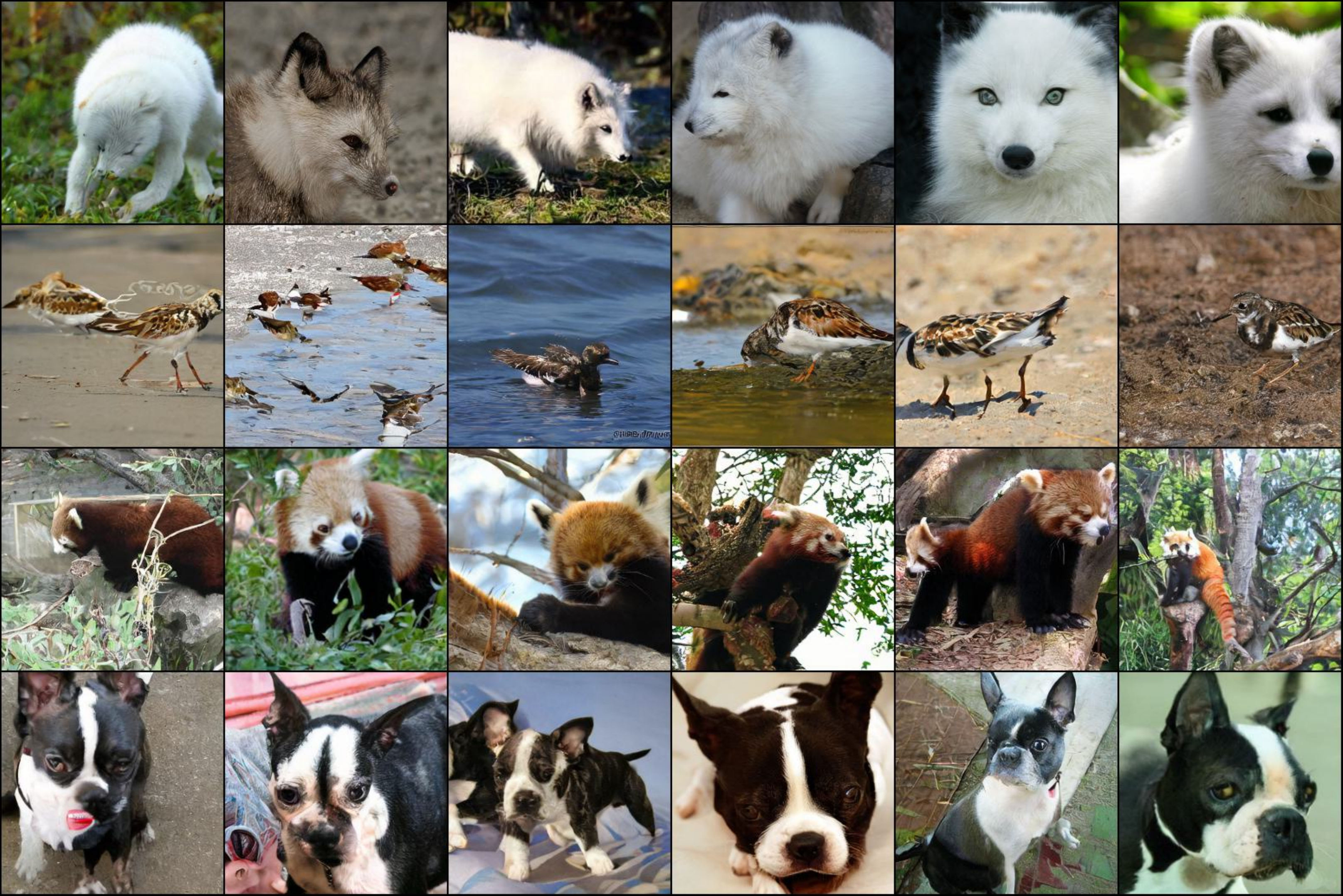}
        \vphantom{g}
        \caption{FP32}
        \label{figure:4_a}
    \end{subfigure}
    \hspace{10pt}
    \begin{subfigure}{0.3\textwidth}
        \centering
        \includegraphics[width=\textwidth]{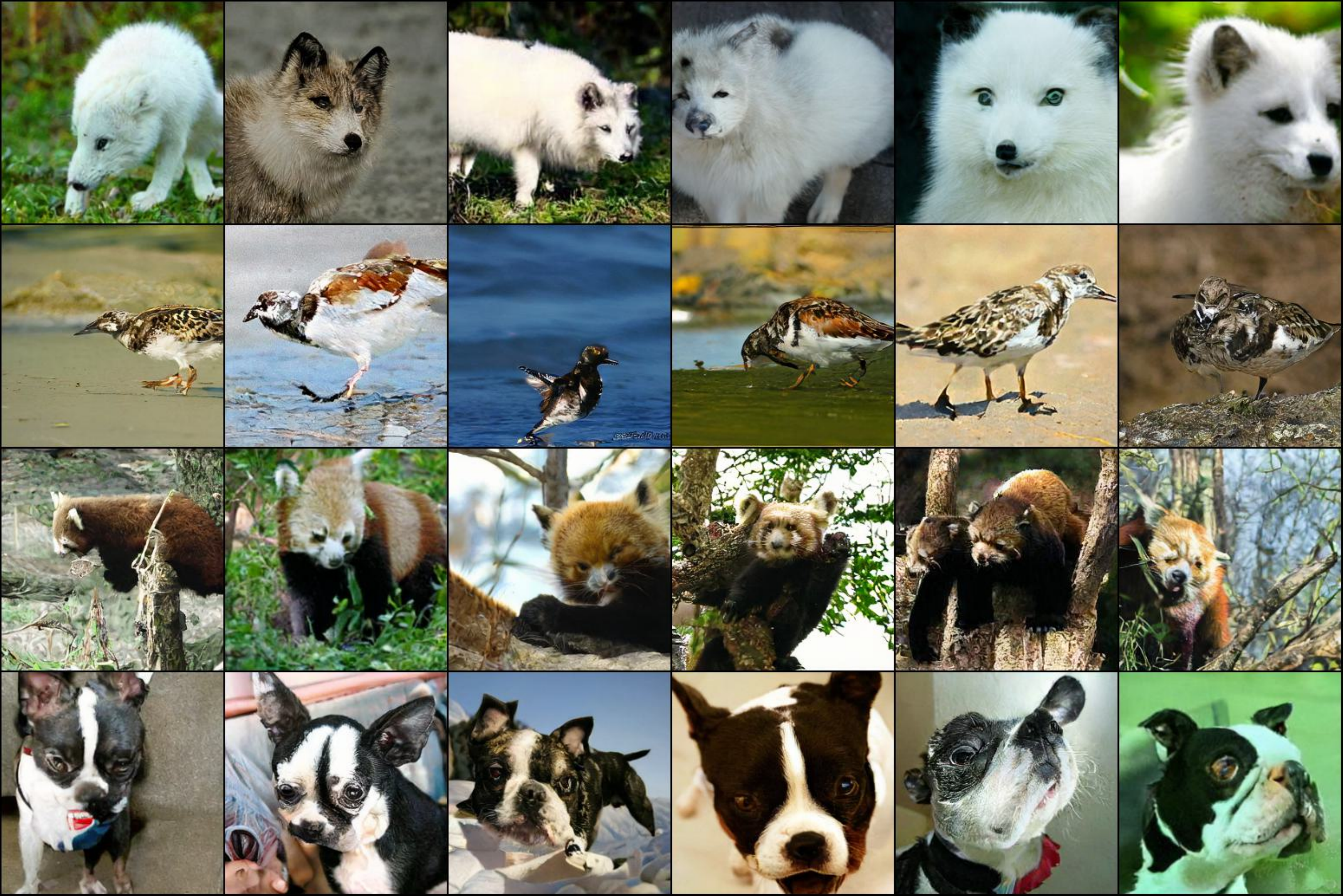}
        \vphantom{g}
        \caption{W4A8}
        \label{figure:4_b}
    \end{subfigure}
    \hspace{10pt}
    \begin{subfigure}{0.3\textwidth}
        \centering
        \includegraphics[width=\textwidth]{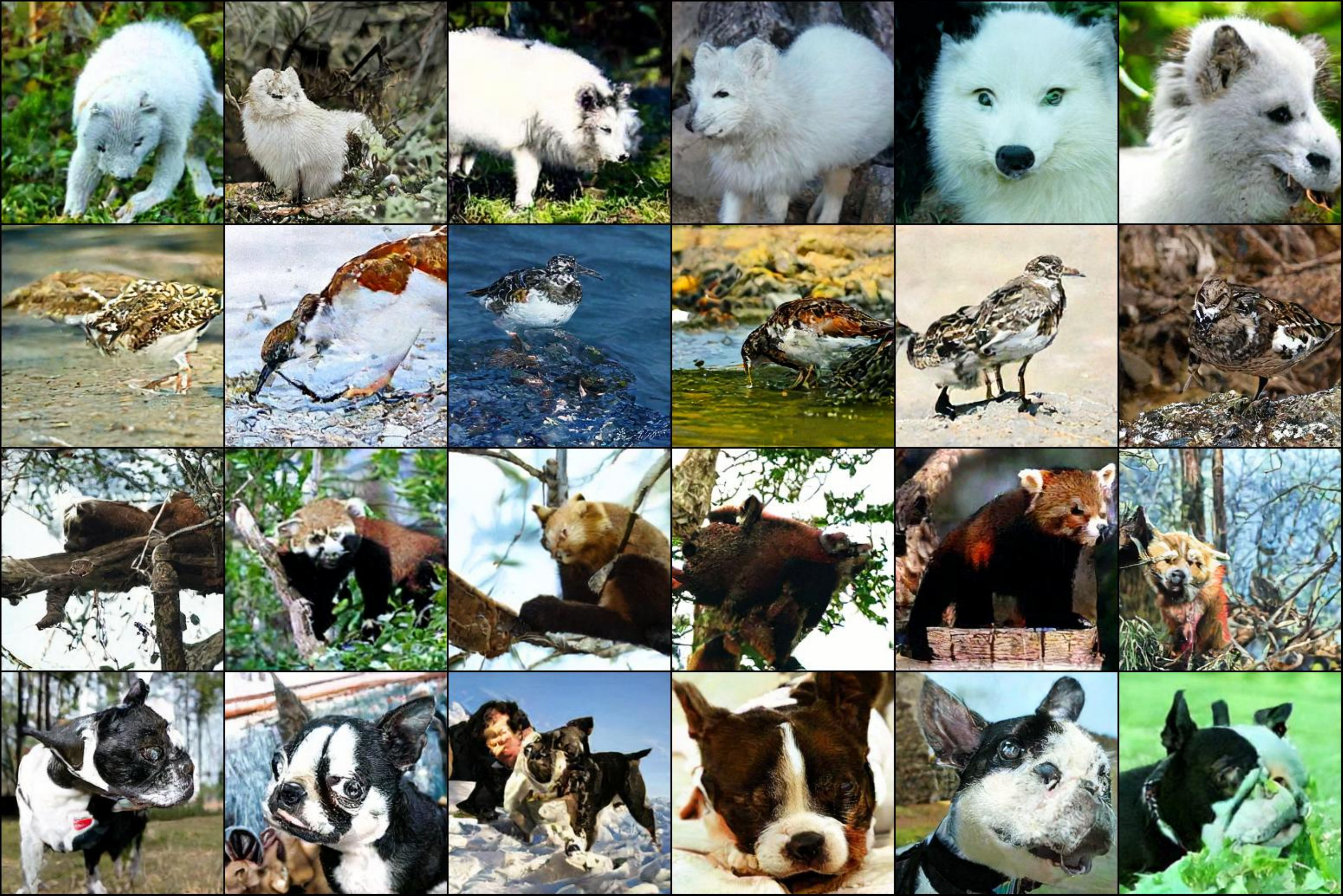}
        \vphantom{g}
        \caption{W4A4}
        \label{figure:4_c}
    \end{subfigure}
    \vspace{10pt}
    \caption{\justifying Visualization of the generative performance of our SA-QLoRA under $W4A8$ and $W4A4$ quantization settings.}
    \label{figure:4}
\end{figure*}
Specifically, on the LSUN-Bedroom dataset, SA-PTQ under $W8A8$ quantization reduces FID and sFID by 0.48 and 2.04, respectively, compared to PTQD. Furthermore, SA-QLoRA under $W4A8$ achieves additional reductions of 1.04 and 1.44. Even under $W4A4$ quantization, SA-QLoRA effectively controls quantization-induced errors, preventing variance explosion and achieving FID and sFID reductions of 4.58 and 4.44, respectively, relative to EfficientDM. On the LSUN-Church dataset, SA-PTQ achieves FID and sFID reductions of 1.22 and 0.46, respectively, over PTQD, while SA-QLoRA further improves these metrics with average reductions of 3.66 in FID and 0.82 in sFID. Experiments on the LSUN dataset further demonstrate that our sampling-aware quantization approach effectively preserves the superior generative performance of high-order samplers under sparse-step sampling, achieving high-fidelity quantization.

\subsection{Text-guided Image Generation}
We evaluate the text-to-image generation task using SD-v1.4 on MS-COCO 512$\times$512, with the results presented in Tab.~\ref{table:3}. Under W8A8 quantization, SA-PTQ achieved consistently optimal metrics, particularly outperforming PTQD by 0.62 in sFID. In the W4A4 setting, SA-QLoRA demonstrated the best performance in terms of FID and CLIP-Score. Further visual results are provided in Appendix.

\begin{table}[h]
\centering
\caption{\justifying Performance evaluation of text-guided image generation on MS-COCO 512 $\times$ 512.}
\footnotesize
\begin{tabular}{cccccc}
\toprule
\textbf{Method} & \textbf{W/A} & TBOPs & \textbf{FID $\downarrow$} & \textbf{sFID $\downarrow$} & \textbf{CLIP-Score. $\uparrow$}  \\ 
\midrule
Q-Diffusion & 8/8 & 51.8 & 13.28 & 20.65 & 0.2904 \\
PTQD & 8/8 & 51.8 & 13.65 & 20.14 & 0.3029 \\
SA-PTQ & 8/8 & 51.8 & \textbf{13.10} & \textbf{19.52} & \textbf{0.3036} \\ 
\cmidrule(lr){1-2} \cmidrule(lr){3-6} 
Q-Diffusion & 4/8 & 25.9 & 14.40 & 21.09 & 0.2875 \\ 
EfficientDM & 4/8 & 25.9 & 13.31 & \textbf{19.92} & 0.3002 \\
SA-QLoRA & 4/8 & 25.9 & \textbf{13.27} & 20.26 & \textbf{0.3017} \\ 
\bottomrule
\end{tabular}%
\label{table:3}
\end{table}

\subsection{Ablation Study}
As shown in Tab.~\ref{table:4}, we conduct ablation studies on the ImageNet 256 $\times$ 256 dataset to validate the effectiveness of the proposed sampling-aware quantization components. Here, MOTAC refers to the Mixed-Order Trajectory Alignment Calibration described in Sec.~\ref{sec:4_2}, while $\mathcal{L}_{MOTA}$ and $\mathcal{L}_{COS}$ represent the two constraints discussed in Sec.~\ref{sec:4_3}. On the PTQ baseline BRECQ, MOTA reduces FID and sFID by 4.1 and 3.83, respectively. Furthermore, MOTA achieves additional reductions of 0.95 in FID and 0.31 in sFID over QLoRA with applied direction alignment constraints. These results demonstrate that our proposed Mixed-Order Trajectory Alignment strategy effectively mitigates sampler performance degradation caused by quantization errors, achieving high-fidelity quantization.
\begin{table}[h]
\centering
\caption{\justifying Ablation study of the sampling-aware quantization components using LDM-4 $(scale = 1.5, step = 20)$ on the ImageNet 256 $\times$ 256.}
\footnotesize
\begin{tabular}{cccccc}
\toprule
\textbf{Method} & \textbf{W/A} & \textbf{IS $\uparrow$} & \textbf{FID $\downarrow$} & \textbf{sFID $\downarrow$} \\ 
\midrule
FP & 32/32 & 174.33 & 9.45 & 8.08 \\ 
\midrule
BRECQ & 8/8 & 112.80 & 14.26 & 13.72 \\ 
+ MOTAC (SA-PTQ) & 8/8 & 120.71 & 10.16 & 9.89 \\ 
\midrule
QLoRA & 4/8 & 132.70 & 9.91 & 8.76 \\ 
+ $\mathcal{L}_{COS}$ & 4/8 & 134.61 & 9.50 & 8.82  \\ 
+ $\mathcal{L}_{MOTA}$ (SA-QLoRA) & 4/8 & 140.56 & 8.55 & 8.51 \\ 
\bottomrule
\end{tabular}%
\label{table:4}
\end{table}

\section{Conclusion}

In this paper, we present a sampling-aware quantization method for diffusion models, designed to achieve high-fidelity dual acceleration. We begin by analyzing the impact of quantization errors on sampling through the lens of sampling acceleration principles, and subsequently introduce a Mixed-Order Trajectory Alignment strategy to quantize a more linear probability flow, thereby mitigating the rapid accumulation of errors in high-speed samplers. Furthermore, we propose two variants, SA-PTQ and SA-QLoRA, to cater to diverse quantization bit-width requirements. Experimental results substantiate that our approach effectively curtails error accumulation during fast sampling, facilitating high-fidelity quantization.


\newpage
\noindent\textbf{Acknowledgements.} This work is funded by National Natural Science Foundation of China (62576305) and Ningbo Youth Science and Technology Innovation Talent Project (2025QL052).
{
    \small
    \bibliographystyle{ieeenat_fullname}
    \bibliography{main}
}
\clearpage
\setcounter{page}{1}
\maketitlesupplementary

\appendix
\section{Theoretical Analysis}
\subsection{High-Order Approximation via Intermediate Point Evaluations in Numerical Integration}\label{appendix:A_1}

For the following ODE \cite{lu2022dpm}:
\begin{equation}
    \frac{\mathrm{d}\mathrm{\mathbf{x}}_t}{\mathrm{d}t} = \alpha \mathrm{\mathbf{x}}_t+\bm{N}(\mathrm{\mathbf{x}}_t, t) , 
     \label{eq:1}
\end{equation}
where $\alpha \in \mathbb{R}$ and $\bm{N}(\mathrm{\mathbf{x}}_t, t) \in \mathbb{R}^D$ is a non-linear function of $\mathrm{\mathbf{x}}_t$. Given an initial value $\mathrm{\mathbf{x}}_t$ at time $t$, for $h>0$, the true solution at time $t + h$ is:
\begin{equation}
    \mathrm{\mathbf{x}}_{t+h} = e^{\alpha h} \mathrm{\mathbf{x}}_t + e^{\alpha h} \int_0^h e^{-\alpha \tau} \bm{N}(\mathrm{\mathbf{x}}_{t+\tau}, t + \tau) \, d\tau.
    \label{eq:2}
\end{equation}
The exponential Runge-Kutta methods \cite{hochbruck2010exponential, hochbruck2005explicit} use some intermediate points to approximate the
integral $\int_0^h e^{-\alpha \tau} \bm{N}(\mathrm{\mathbf{x}}_{t+\tau}, t + \tau) \, d\tau$. Accordingly, DPM-Solver adopts this method to compute the analogous integral in Eqn.~\eqref{eq:3} with $\alpha = 1$ and $\bm{N}=\boldsymbol{\epsilon}_{\theta}$:
\begin{equation}
    \mathrm{\mathbf{x}}_{\lambda_t + h} = \frac{\alpha_{\lambda_t + h}}{\alpha_{\lambda_t}} \mathrm{\mathbf{x}}_{\lambda_t} + \alpha
    _{\lambda_t + h} \int _{\lambda_t} ^{\lambda_t + h} e^{-\lambda} \boldsymbol{\epsilon}_{\theta}(\mathrm{\mathbf{x}}_{\lambda}, \lambda) \, \mathrm{d}\lambda
    \label{eq:3}
\end{equation}
This is equivalent to approximating the continuous integral using a higher-order Taylor expansion of $\mathrm{\mathbf{x}}(\lambda + h)$ at $\lambda = \lambda_t$. For an in-depth theoretical foundation of numerical methods, refer to \cite{hochbruck2010exponential, hochbruck2005explicit}. Here, we present a concise derivation of the expansion corresponding to the second-order Runge-Kutta method.

First, we make the following assumptions to ensure the applicability of the $k$-th order Taylor expansion:
 
\noindent\textbf{Assumption \#1:} The total derivatives of $\boldsymbol{\epsilon}_{\theta}(\mathrm{\mathbf{x}}_{\lambda}, \lambda)$, denoted as $\frac{\partial ^j \boldsymbol{\epsilon}_{\theta}(\mathrm{\mathbf{x}}_{\lambda}, \lambda)}{\partial \mathrm{\mathbf{x}}_\lambda^j}$ and $\frac{\partial ^j \boldsymbol{\epsilon}_{\theta}(\mathrm{\mathbf{x}}_{\lambda}, \lambda)}{\partial \lambda^j}$, exist and are continuous for all $0 \leq j \leq k+1$.

\noindent\textbf{Assumption \#2:}\label{assumption:2} The step size $h = \lambda_t - \lambda_s$ satisfies $h = \mathcal{O}(\frac{1}{N})$, where $N$ is the number of integration steps, ensuring the step size is sufficiently small.

\noindent\textbf{Analysis.} In denoising diffusion, for the simplified probability flow integral:
\begin{equation}
    \mathrm{\mathbf{x}}_{\lambda_t + h} = \mathrm{\mathbf{x}}_{\lambda_t} + \int _{\lambda_t}^{\lambda_t + h}\boldsymbol{\epsilon}_{\theta}(\mathrm{\mathbf{x}}_{\lambda}, \lambda)\mathrm{d}\lambda,
    \label{eq:4}
\end{equation}
the general form of the second-order Runge-Kutta method is:
\begin{equation}
\begin{cases}
k_1 = \bm{\epsilon}_{\theta}(\mathrm{\mathbf{x}}_{\lambda_t}, \lambda_t), \\
k_2 = \bm{\epsilon}_{\theta}(\mathrm{\mathbf{x}}_{\lambda_t} + bhk_1, \lambda_t + ah), \\
\mathrm{\mathbf{x}}_{\lambda_t + h} = \mathrm{\mathbf{x}}_{\lambda_t} + h\left[\left(1 - \frac{1}{2a}\right)k_1 + \frac{1}{2a}k_2\right].
\end{cases}
\label{eq:5}
\end{equation}

For the classical midpoint method, taking $a = b = \frac{1}{2}$, we have:
\begin{equation}
\begin{cases}
k_1 = \bm{\epsilon}_{\theta}(\mathrm{\mathbf{x}}_{\lambda_t}, \lambda_t), \\
k_2 = \bm{\epsilon}_{\theta}(\mathrm{\mathbf{x}}_{\lambda_t} + \frac{h}{2}k_1, \lambda_t + \frac{h}{2}), \\
\mathrm{\mathbf{x}}_{\lambda_t + h} = \mathrm{\mathbf{x}}_{\lambda_t} + hk_2.
\end{cases}
\label{eq:6}
\end{equation}
Then, for $k_2$, perform a first-order Taylor expansion of 
$\bm{\epsilon}_{\theta}(\mathrm{\mathbf{x}}_{\lambda}, \lambda)$ at $(\mathrm{\mathbf{x}}_{\lambda_t}, \lambda_t)$, yielding:
\begin{footnotesize}
\begin{align}
    k_2 &= \bm{\epsilon}_{\theta}(\mathrm{\mathbf{x}}_{\lambda_t} + \frac{h}{2}k_1, \lambda_t + \frac{h}{2}) \notag \\
    &= \bm{\epsilon}_{\theta}(\mathrm{\mathbf{x}}_{\lambda_t}, \lambda_t) + \left. \frac{\partial \boldsymbol{\epsilon}_{\theta}(\mathrm{\mathbf{x}}_{\lambda}, \lambda)}{\partial \lambda} \right|_{(\mathrm{\mathbf{x}}_{\lambda_t}, \lambda_t)} \cdot \frac{h}{2} + \mathcal{O}(h^2) \notag \\
    &+ \left. \frac{\partial \boldsymbol{\epsilon}_{\theta}(\mathrm{\mathbf{x}}_{\lambda}, \lambda)}{\partial \mathrm{\mathbf{x}}_{\lambda}} \right|_{(\mathrm{\mathbf{x}}_{\lambda_t}, \lambda_t)} \cdot \frac{h}{2} k_1 \notag \\
    &= \bm{\epsilon}_{\theta}(\mathrm{\mathbf{x}}_{\lambda_t}, \lambda_t) + \frac{h}{2} \frac{\partial \boldsymbol{\epsilon}_{\theta}}{\partial \lambda}(\mathrm{\mathbf{x}}_{\lambda_t}, \lambda_t) \notag + \frac{h}{2} \frac{\partial \bm{\epsilon}_{\theta}}{\partial \mathrm{\mathbf{x}}_{\lambda}}(\mathrm{\mathbf{x}}_{\lambda_t}, \lambda_t) \bm{\epsilon}_{\theta}(\mathrm{\mathbf{x}}_{\lambda_t}, \lambda_t) \notag \\ 
    &+ \mathcal{O}(h^2)
    \label{eq:7}
\end{align}
\end{footnotesize}
Substituting $k_2$ into Eqn.~\eqref{eq:5}, we obtain:
\begin{align}
\mathbf{x}_{\lambda_t + h} &= \mathbf{x}_{\lambda_t} + h \big[ 
\bm{\epsilon}_{\theta}(\mathbf{x}_{\lambda_t}, \lambda_t) 
+  \frac{h}{2}\frac{\partial \bm{\epsilon}_{\theta}}{\partial \lambda}(\mathbf{x}_{\lambda_t}, \lambda_t) \notag \\
&\quad + \frac{h}{2} \frac{\partial \bm{\epsilon}_{\theta}}{\partial \mathrm{\mathbf{x}}_{\lambda}}(\mathbf{x}_{\lambda_t}, \lambda_t) \boldsymbol{\epsilon}_{\theta}(\mathrm{\mathbf{x}}_{\lambda_t}, \lambda_t)
+ \mathcal{O}(h^2) \big] \notag \\
&= \mathrm{\mathbf{x}}_{\lambda_t} + h \boldsymbol{\epsilon}_{\theta}(\mathrm{\mathbf{x}}_{\lambda_t}, \lambda_t) + \frac{h^2}{2} \big[ \frac{\partial \boldsymbol{\epsilon}_{\theta}}{\partial \lambda}(\mathrm{\mathbf{x}}_{\lambda_t}, \lambda_t) \notag \\
&+ \frac{\partial \boldsymbol{\epsilon}_{\theta}}{\partial \mathrm{\mathbf{x}}_{\lambda_t}}(\mathrm{\mathbf{x}}_{\lambda_t}, \lambda_t) \boldsymbol{\epsilon}_{\theta}(\mathrm{\mathbf{x}}_{\lambda_t}, \lambda_t)
\big] + \mathcal{O}(h^3)
\label{eq:8}
\end{align}
Thus, this is equivalent to the second-order Taylor expansion of $\mathrm{\mathbf{x}}(\lambda + h)$ at $\lambda = \lambda_t$:
\begin{align}
    \mathrm{\mathbf{x}}(\lambda_t+h) &= \mathrm{\mathbf{x}}(\lambda_t) + h \mathrm{\mathbf{x}}'(\lambda_t) + \frac{h^2}{2}\mathrm{\mathbf{x''}}(\lambda_t) + \mathcal{O}(h^3) \notag \\
    &= \mathrm{\mathbf{x}}(\lambda_t) + h \boldsymbol{\epsilon}_{\theta}(\mathrm{\mathbf{x}}_{\lambda_t}, \lambda_t) + \frac{h^2}{2} \big[ \frac{\partial \boldsymbol{\epsilon}_{\theta}}{\partial \lambda}(\mathrm{\mathbf{x}}_{\lambda_t}, \lambda_t) \notag \\
    &+ \frac{\partial \boldsymbol{\epsilon}_{\theta}}{\partial \mathrm{\mathbf{x}}_{\lambda}}(\mathrm{\mathbf{x}}_{\lambda_t}, \lambda_t) \boldsymbol{\epsilon}_{\theta}(\mathrm{\mathbf{x}}_{\lambda_t}, \lambda_t)
    \big] + \mathcal{O}(h^3)
    \label{eq:9}
\end{align}
Moreover, from Eqn.~\eqref{eq:7}, it can be observed that the evaluation $\boldsymbol{\epsilon}_{\theta}(\mathrm{\mathbf{x}}_{\lambda_t} + bhk_1, \lambda_t + ah)$ at the midpoint $(\mathrm{\mathbf{x}}_{\lambda_t} + bhk_1, \lambda_t + ah)$ contributes derivative information $\boldsymbol{\epsilon}_{\theta}^{(1)}(\mathrm{\mathbf{x}}_{\lambda_t}, \lambda_t)$ to the second-order Taylor expansion in Eqn.~\eqref{eq:9}.
\subsection{Quantization Error Analysis in Fast Sampling of Quantized Diffusion Models}\label{appendix:A_2}
To compute the numerical integration over the interval $(\lambda_s, \lambda_t)$ corresponding to Eqn.~\eqref{eq:3}, the sampler approximates the sampling direction of the continuous equation by solving the higher-order expansion of $\boldsymbol{\epsilon}_{\theta}(\mathrm{\mathbf{x}}_{\lambda}, \lambda)$ at $(\mathrm{\mathbf{x}}_{\lambda_s}, \lambda_s)$:

\begin{footnotesize}
\begin{align}
\mathbf{x}_t & = \frac{\alpha_t}{\alpha_s} \mathbf{x}_s -\alpha_t \sum_{n=0}^{k-1} \boldsymbol{\epsilon}_{\theta}^{(n)}(\mathbf{x}_{\lambda_s}, \lambda_s) \int_{\lambda_s}^{\lambda_t} 
e^{-\lambda} \cdot \frac{(\lambda - \lambda_s)^n}{n!}\, \mathrm{d}\lambda \notag \\
& + \mathcal{O}((\lambda_t-\lambda_s)^{k+1}),
\label{eq:10}
\end{align}
\end{footnotesize}
\noindent where $\boldsymbol{\epsilon}_{\theta}^{(n)}(\mathbf{x}_{\lambda_s}, \lambda_s) = \frac{\mathrm{d}^n\boldsymbol{\epsilon}_{\theta}^{(n)}(\mathbf{x}_{\lambda}, \lambda)}{\mathrm{d}{\lambda}^n} $ is the $n$-th order total derivative of w.r.t. $\lambda$. However, the quantized model $\boldsymbol{\hat{\epsilon}}_{\theta}$ introduces quantization errors $\Delta\boldsymbol{\epsilon}_{\theta}$, transforming the integral into:

\begin{footnotesize}
\begin{align}
\mathbf{x}_t &= \frac{\alpha_t}{\alpha_s} \mathbf{x}_s - \alpha_t \sum_{n=0}^{k-1} \boldsymbol{\epsilon}_{\theta}^{(n)}(\mathbf{x}_{\lambda_s}, \lambda_s) \int_{\lambda_s}^{\lambda_t} 
e^{-\lambda} \cdot \frac{(\lambda - \lambda_s)^n}{n!}\, \mathrm{d}\lambda \notag \\
&+ \sum_{n=0}^{k-1} \Delta\boldsymbol{\epsilon}_{\theta}^{(n)}(\mathbf{x}_{\lambda_s}, \lambda_s) \int_{\lambda_s}^{\lambda_t} 
e^{-\lambda} \cdot \frac{(\lambda - \lambda_s)^n}{n!}\, \mathrm{d}\lambda \notag \\
&+ \mathcal{O}((\lambda_t-\lambda_s)^{k+1}) \,
\label{eq:11}
\end{align}
\end{footnotesize}

\noindent Next, we denote $\Delta_{quant} = \sum_{n=0}^{k-1} \Delta\boldsymbol{\epsilon}_{\theta}^{(n)}(\mathbf{x}_{\lambda_s}, \lambda_s) \int_{\lambda_s}^{\lambda_t} 
e^{-\lambda} \cdot \frac{(\lambda - \lambda_s)^n}{n!}\, \mathrm{d}\lambda$ as the quantization cumulative error, $\Delta_{disc}$ as the discretization truncation error, and proceed to analyze the upper bound of the quantization cumulative error.

\noindent \textbf{Analysis.} First, we compute the integral:
\begin{equation}
    I = \int_{\lambda_s}^{\lambda_t}e^{-\lambda} \cdot \frac{(\lambda-\lambda_s)^n}{n!} \mathrm{d}\lambda 
    \label{eq:12}
\end{equation}
Define $u = \lambda - \lambda_s$, which implies $\lambda = u + \lambda_s, \mathrm{d}\lambda = \mathrm{d}u$. Substituting these into Eqn.~\eqref{eq:11}, we obtain:
\begin{align}
    I &= \int_0^{\lambda_t - \lambda_s} e^{-(u+\lambda_s)} \cdot \frac{u^n}{n!} \mathrm{d}u \notag \\
    &= \frac{e^{-\lambda_s}}{n!}\int_0^{\lambda_t - \lambda_s} e^{-u} \cdot u^n \mathrm{d}u \notag \\
    &= \frac{e^{-\lambda_s}}{n!} \cdot \gamma(n+1, \lambda_t - \lambda_s),
    \label{eq:13}
\end{align}
where $\gamma(\cdot, \cdot)$ denotes the lower incomplete Gamma function. According to \textbf{Assumption \#2}, the step size h is small, and $s < t$, thus $(\lambda_t - \lambda_s) \to 0^+$. Under this condition, $\gamma(\cdot, \cdot)$ can be approximated as:
\begin{equation}
    \gamma(n+1, \lambda_t - \lambda_s) \approx \frac{(\lambda_t - \lambda_s)^{n+1}}{n+1},
    \label{eq:14}
\end{equation}
thus, the integral $I$ simplifies to:
\begin{equation}
    I = \frac{e^{-\lambda_s} \cdot (\lambda_t - \lambda_s)^{n+1}}{(n+1)!}
    \label{eq:15}
\end{equation}
Considering the convergence of the quantization algorithm, we assume that the quantization error is bounded:
\begin{equation}
    \left| \Delta \epsilon_{\theta}^{(n)} \left( \mathbf{x}_{\lambda_s}, \lambda_s \right) \right| \leq \delta_n
    \label{eq:16}
\end{equation}
\begin{equation}
    \delta = \max_{i \in \mathcal{I}} \delta_i, \quad \mathcal{I} = \{1, 2, \ldots, n\}
    \label{eq:17}
\end{equation}
thus, according to the triangle inequality, we have:
\begin{align}
    \left| \Delta_{quant} \right| &= \left| \Sigma_{n=0}^{k-1} \Delta\boldsymbol{\epsilon_{\theta}}^{(n)}(\mathrm{\mathbf{x}}_{\lambda_s}, \lambda_s) \int_{\lambda_s}^{\lambda_t}e^{-\lambda}\frac{(\lambda-\lambda_s)^n}{n!}\mathrm{d}\lambda \right| \notag \\
    &\leq \Sigma_{n=0}^{k-1}\left|\Delta\boldsymbol{\epsilon}_{\theta}^{(n)}(\mathrm{\mathbf{x}}_{\lambda_s}, \lambda_s) \right|\cdot \left| \int_{\lambda_s}^{\lambda_t}e^{-\lambda}\frac{(\lambda-\lambda_s)^n}{n!}\mathrm{d}\lambda \right| \notag \\
    &\leq \Sigma_{n=0}^{k-1} \, \delta \, \cdot \, e^{-\lambda_s} \cdot \, \frac{(\lambda_t - \lambda_s)^{n+1}}{(n+1)!}
    \label{eq:18}
\end{align}
Define the $n$-th order derivative error $a_n = \delta \cdot e^{-\lambda_s} \cdot \frac{(\lambda_t - \lambda_s)^{n+1}}{(n+1)!}$, then, the cumulative quantization error satisfies  $\left| \Delta_{quant} \right| \leq \Sigma_{n=0}^{k-1} \, a_n$, where the ratio of successive terms is given by:
\begin{align}
    \frac{a_{n}}{a_{n-1}} &= \frac{\delta \, \cdot \, e^{-\lambda_s} \, \cdot \, \frac{(\lambda_t - \lambda_s)^{n+1}}{(n+1)!}}{\delta \, \cdot \, e^{-\lambda_s} \, \cdot \, \frac{(\lambda_t - \lambda_s)^n}{n!}} \notag \\
    &= \frac{\lambda_t - \lambda_s}{n+1}
    \label{eq:19}
\end{align}
According to the previous assumption that $\lambda_t - \lambda_s \ll 1$, it follows that $a_n \ll a_{n-1}$, indicating that $a_n$ decreases rapidly as the order $n$ increases. Consequently, the error upper bound is estimated as:
\begin{equation}
    \mathcal{L}_{\Delta_{quant}} = \mathcal{O}(\delta \, \cdot \, e^{-\lambda_s} \, (\lambda_t - \lambda_s))
    \label{eq:20}
\end{equation}
\begin{align}
\mathcal{L}_{\Delta} &= \mathcal{L}_{\Delta_{quant}} + \mathcal{L}_{\Delta_{disc}} \notag \\
&= \mathcal{O}(\delta \cdot e^{-\lambda_s} \cdot (\lambda_t - \lambda_s)) + \mathcal{O}((\lambda_t - \lambda_s)^{k+1})
\label{eq:21}   
\end{align}

\section{Related Work on Sampling Acceleration} \label{appendix:B}
Advanced accelerated sampling algorithms approximate the continuous sampling equations using high-precision numerical integration methods, minimizing truncation errors introduced by discretization. This enables larger sampling step sizes, thus reducing the number of required sampling steps while maintaining accuracy. DDIM \cite{song2020denoising} achieves non-Markovian skip-step sampling by aligning marginal probability distributions, essentially leveraging a first-order Euler discretization to approximate the solution of the neural ODE. DPM-solver \cite{lu2022dpm, lu2022dpmplus} performs a high-order expansion of the noise estimation network at discrete steps to approximate the sampling direction of the corresponding analytical integral. AMED-Solver \cite{zhou2024fast} utilizes an embedded network to estimate the direction and step size of the subsequent step, incurring a minor increase in computational overhead during inference. PNMD \cite{liu2022pseudo} introduces a pseudo-numerical solving approach, further enhancing the accuracy of traditional numerical solvers. Complementing these trajectory-level acceleration strategies, parallel and sparse decoding techniques offer orthogonal efficiency gains: dParallel \cite{chen2025dparallel} unlocks the inherent parallelism of diffusion language models for fast sampling, while SparseD \cite{wang2026sparsed} accelerates inference by introducing sparse attention.

\section{Experimental Details and Results}
\subsection{Quantization Settings.} \label{appendix:C_1}
To comprehensively evaluate the proposed sampling-aware quantization framework, we conduct experiments under three quantization configurations: $W8A8$, $W4A8$, and $W4A4$. For the $W8A8$ setting, we assess the performance of the proposed SA-PTQ, while for $W4A8$ and $W4A4$ configurations, we employ SA-QLoRA for evaluation. Consistent with prior work, the first and last layers are quantized to 8 bits, with all other layers quantized to the target bit-width. Regarding data calibration, SA-PTQ utilizes the proposed dual-order trajectory sampling to gather the calibration dataset, whereas SA-QLoRA first collects the full-precision first-order trajectory to initialize the quantization parameters. 

\subsection{SA-QLoRA Finetuning Details} \label{appendix:C_2}
In SA-QLoRA fine-tuning, we set the batch size to 4, the adapter rank to 32, and the number of training epochs to 160. To further enhance the alignment of sparse-step sampling trajectories, we design a mixstep progressive LoRA strategy. The basic QLoRA strategy fixes the sampling steps and aligns the full-precision and quantized outputs at each step of the sampler. In contrast, the mixstep progressive LoRA strategy iterates over a list of sampling steps set to steps = [100, 50, 20]. For each cycle, the sampler updates to the current steps[i] value, and the alignment is performed at each sampling step between $\boldsymbol{\epsilon}_{\theta}(\mathrm{\mathbf{x}}_{t_i}, t_i)$ and $\boldsymbol{\hat{\epsilon}}_{\theta}(\mathrm{\mathbf{x}}_{s_i}, s_i)$, where $(\mathrm{\mathbf{x}}_{t_i}, t_i)$ denotes first-order sampling step and $(\mathrm{\mathbf{x}}_{s_i}, s_i)$ denotes intermediate step in second-order sampling.

\subsection{Unconditional Image Generation on LSUN-Churches 256$\times$256}

\begin{table}[hb]
\centering
\caption{\justifying Performance comparisons of unconditional image generation on LSUN-Church 256 $\times$ 256 using the LDM-8 model.}
\footnotesize
\begin{tabular}{cccccc}
\toprule
\textbf{Method} & \textbf{W/A} & \textbf{FID $\downarrow$} & \textbf{sFID $\downarrow$} & \textbf{Prec. $\uparrow$} & \textbf{Rec. $\uparrow$}  \\ 
\midrule
FP & 32/32 & 7.26 & 13.75 & 61.50$\%$ & 50.72$\%$ \\ 
\midrule
PTQD & 8/8 & 11.87 & 12.97 & 56.57$\%$ & 54.15$\%$ \\ 
SA-PTQ & 8/8 & \textbf{10.65} & \textbf{12.51} & \textbf{56.86}$\%$ & \textbf{54.54}$\%$  \\ 
\cmidrule(lr){1-2} \cmidrule(lr){3-6} 
PTQD & 4/8 & 12.96 & 17.81 & 50.23$\%$ & 52.80$\%$ \\ 
EfficientDM & 4/8 & 11.86 & 15.64 & 52.27$\%$ & \textbf{53.78$\%$} \\ 
SA-QLoRA & 4/8 & \textbf{10.07} & \textbf{15.11} & \textbf{54.15$\%$} & 53.68$\%$  \\ 
\cmidrule(lr){1-2} \cmidrule(lr){3-6} 
PTQD & 4/4 & - & - & - & - \\ 
EfficientDM & 4/4 & 23.42 & 20.15 & 46.02$\%$ & \textbf{45.63$\%$} \\ 
SA-QLoRA & 4/4 & \textbf{17.89} & \textbf{19.04} & \textbf{48.95}$\%$ & 43.07$\%$  \\ 
\bottomrule
\end{tabular}%
\label{table:5}
\end{table}

\clearpage
\onecolumn
\section{Additional Visual Results}
\subsection{Visualization of Multi-Order Trajectories}
\begin{figure*}[h!]
    \centering
    \begin{subfigure}{0.4\textwidth} 
        \centering
        \includegraphics[width=\textwidth]{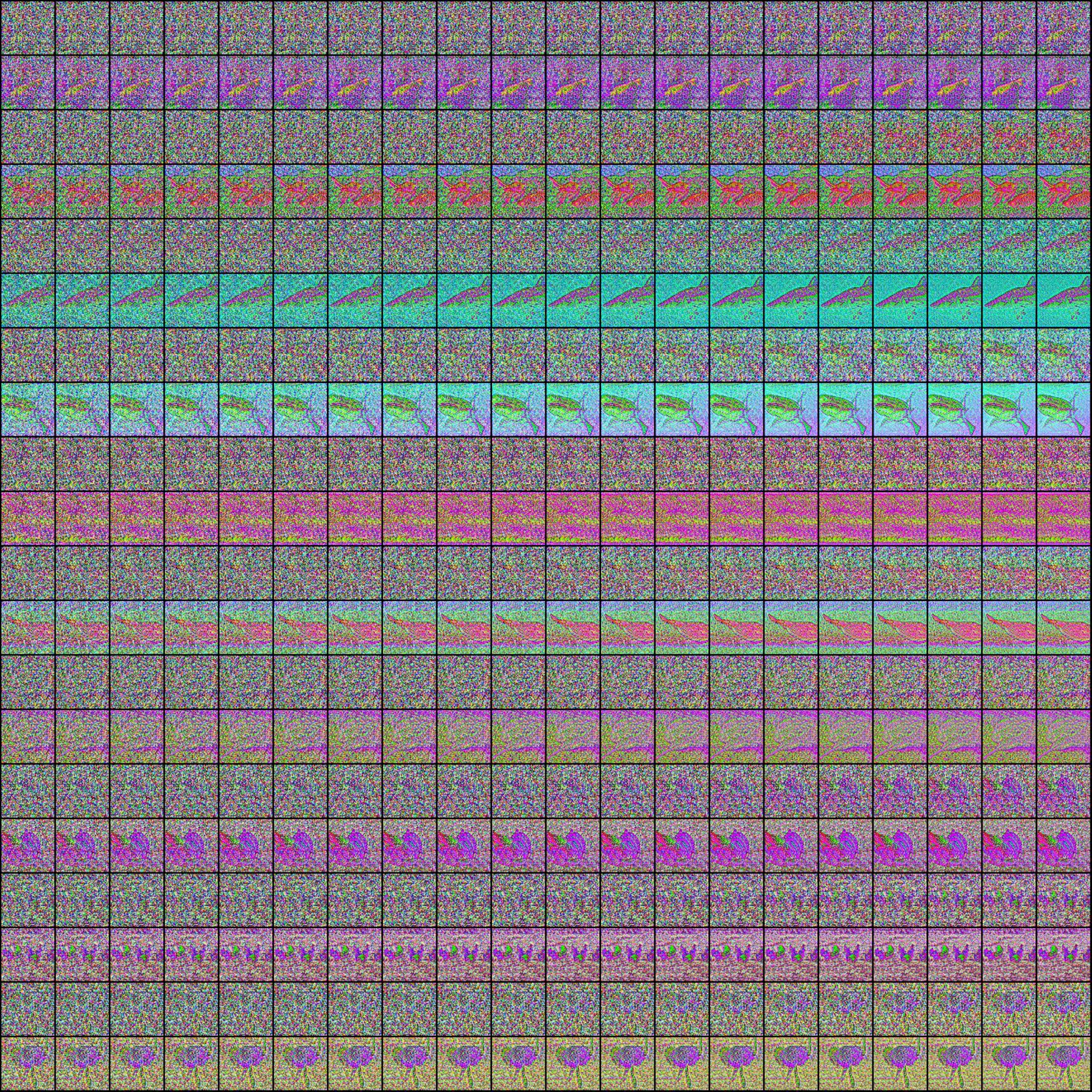}
        \vspace{0pt}
        \caption{}
        \label{fig:traj2}
    \end{subfigure}
    \hspace{20pt} 
    \begin{subfigure}{0.4\textwidth} 
        \centering
        \includegraphics[width=\textwidth]{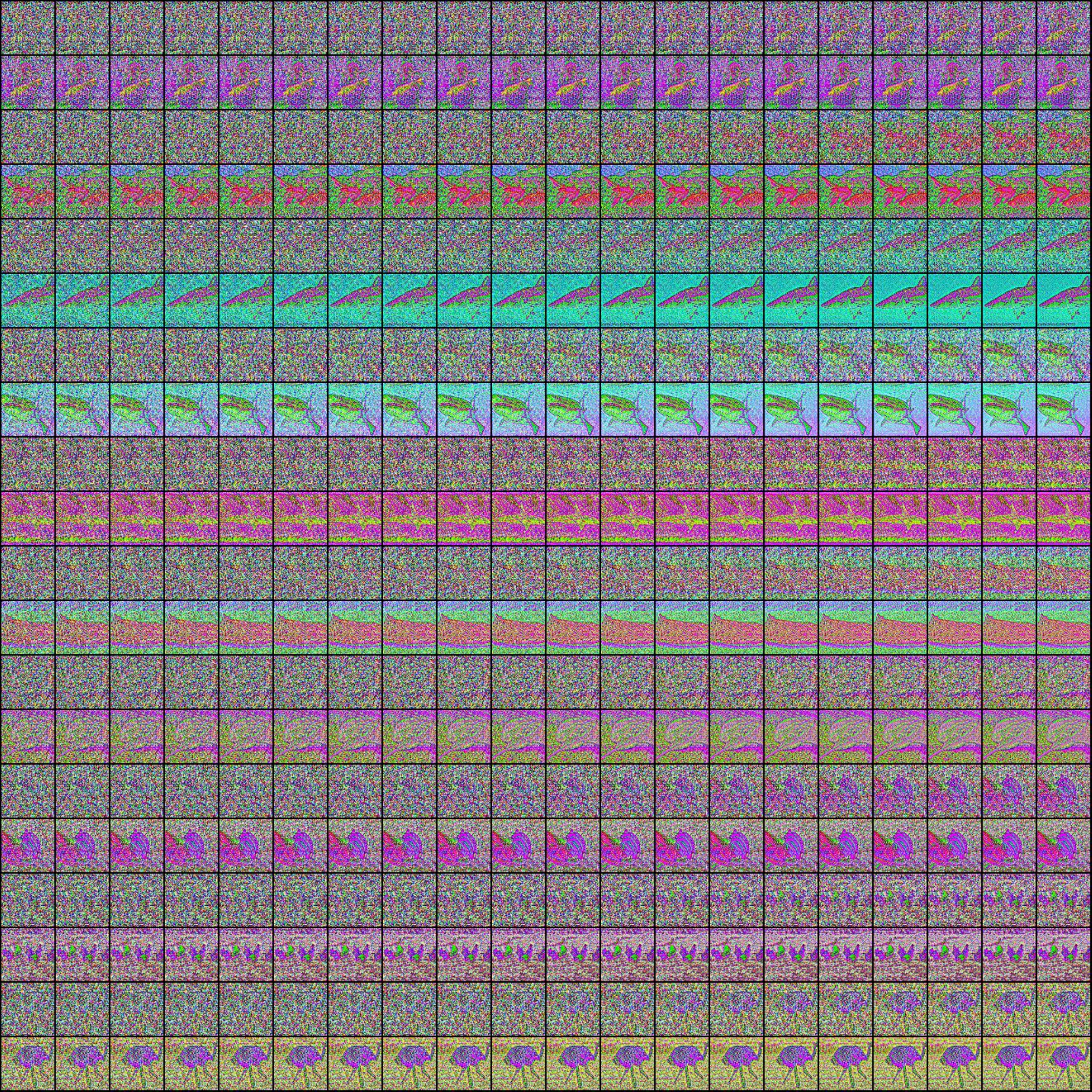}
        \vspace{0pt}
        \caption{}
        \label{fig:traj1}
    \end{subfigure}
    \vspace{2pt}
    \caption{\justifying Latent space feature trajectories of LDM4 under 20-step sampling on the ImageNet 256$\times$256 dataset. (a) Feature trajectories sampled using DPM-Solver-1. (b) Intermediate-step feature trajectories sampled using DPM-Solver-2.}
    \label{fig:trajectory}
\end{figure*}
\subsection{Visual Comparison Across Quantization Algorithms} \label{appendix:d}

\begin{figure*}[ht]
    \centering
    \includegraphics[width=0.75\textwidth]{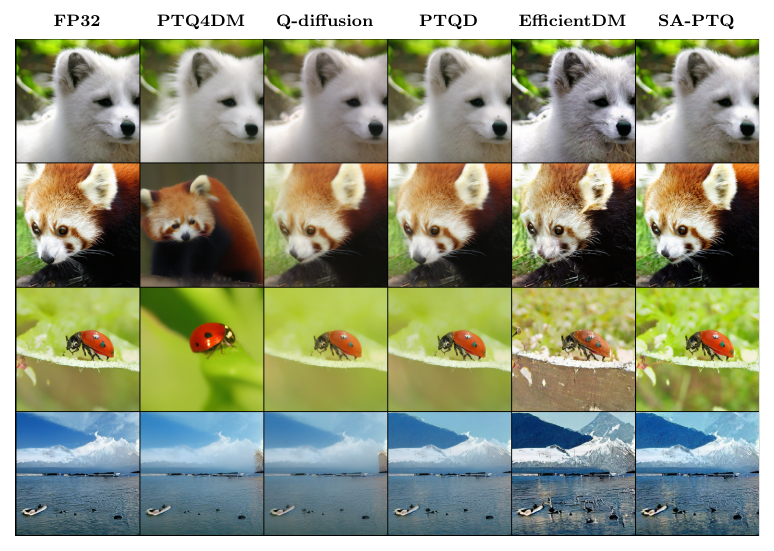}
    \caption{\justifying Comparison of generative performance on the ImageNet 256$\times$256 dataset with 20-step sampling among the full-precision LDM4 and its W8A8 quantized counterparts using PTQ4DM, Q-diffusion, PTQD, EfficientDM, and our proposed SA-LoRA. (Revised version of the main figure in the main text, supplemented with the names of the applied quantization algorithms.)}
    \label{fig:main_supp}
\end{figure*}

\begin{figure*}[ht]
    \centering
    \includegraphics[width=0.75\textwidth]{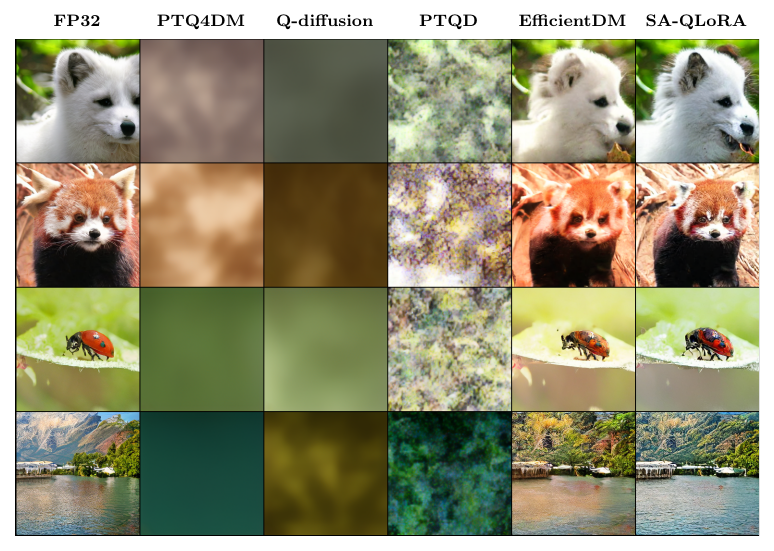}
    \caption{\justifying Comparison of generative performance on the ImageNet 256$\times$256 dataset with 20-step sampling among the full-precision LDM4 and its W4A4 quantized counterparts using PTQ4DM, Q-diffusion, PTQD, EfficientDM, and our proposed SA-LoRA. (Revised version of the main figure in the main text, supplemented with the names of the applied quantization algorithms.)}
    \label{fig:main_supp}
\end{figure*}

\begin{figure*}[h!]
    \centering
    \begin{subfigure}{0.45\textwidth} 
        \centering
        \includegraphics[width=\textwidth]{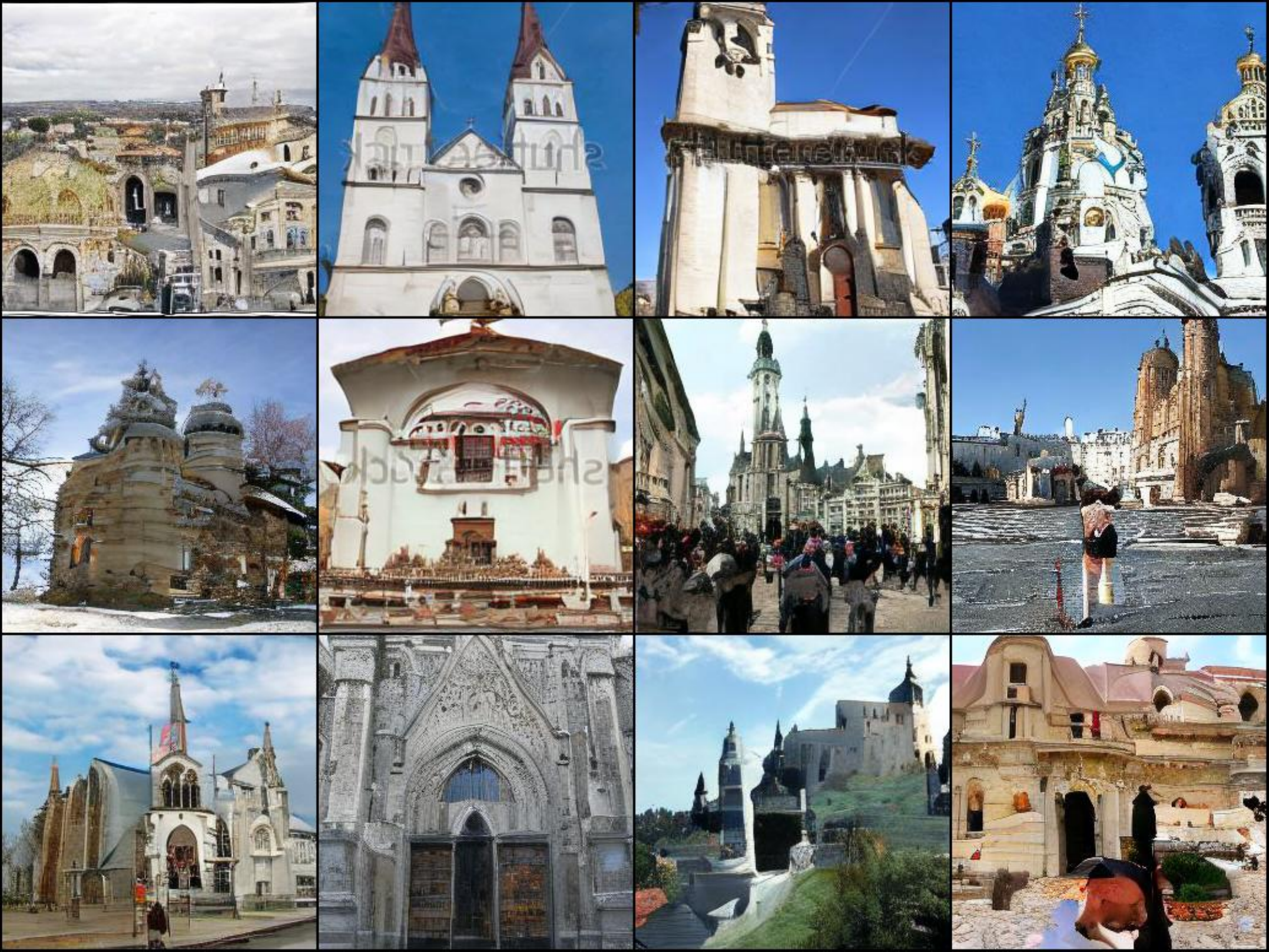}
        \vspace{0pt}
        \caption{FP32}
    \end{subfigure}
    \begin{subfigure}{0.45\textwidth} 
        \centering
        \includegraphics[width=\textwidth]{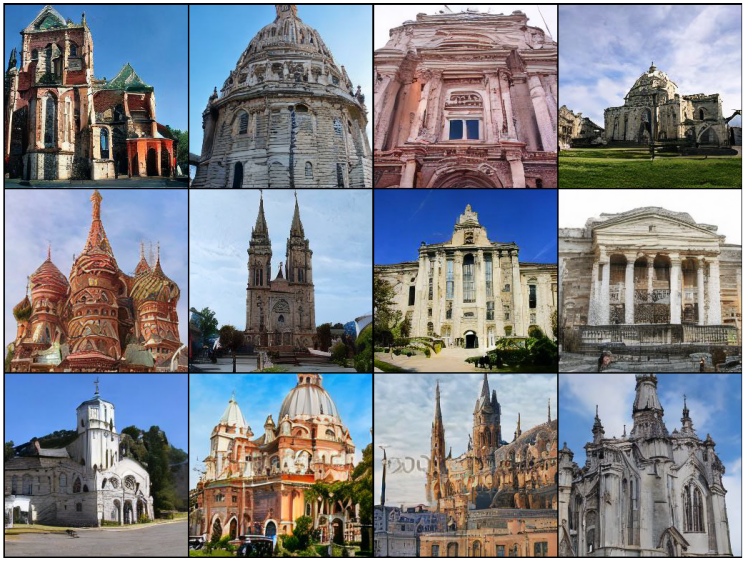}
        \vspace{0pt}
        \caption{SA-QLoRA $\left[W4A8\right]$}
    \end{subfigure}
    \caption{\justifying Comparison of generative performance between the full-precision LDM8 and its W4A8 quantized counterpart, utilizing our proposed SA-QLoRA, on the LSUN-Church 256$\times$256 dataset under 50-step sampling.}
    \label{fig:trajectory}
\end{figure*}

\begin{figure*}[h!]
    \centering
    \begin{subfigure}{0.45\textwidth} 
        \centering
        \includegraphics[width=\textwidth]{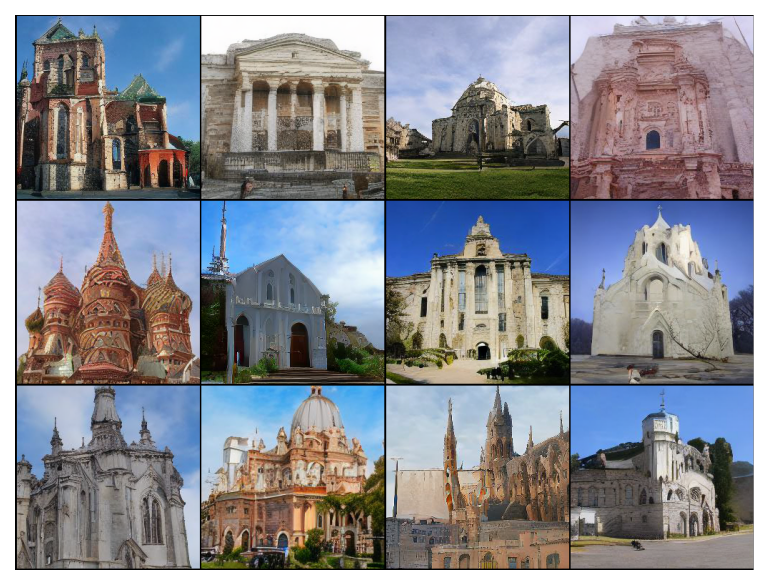}
        \vspace{0pt}
        \caption{PTQD $\left[W4A8\right]$}
    \end{subfigure}
    \begin{subfigure}{0.45\textwidth} 
        \centering
        \includegraphics[width=\textwidth]{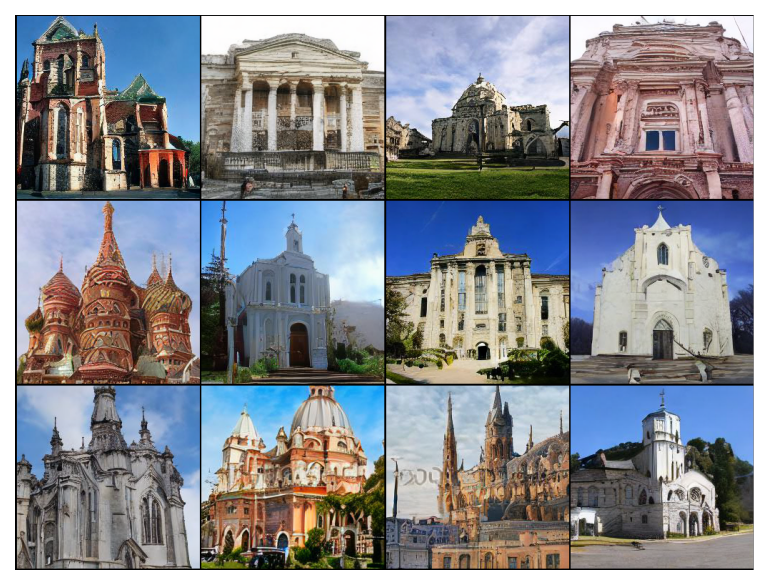}
        \vspace{0pt}
        \caption{SA-QLoRA $\left[W4A8\right]$}
    \end{subfigure}
    \caption{\justifying Generative performance comparison of W4A8 quantized LDM8 models, employing PTQD and our proposed SA-QLoRA, on the LSUN-Church 256$\times$256 dataset with 50-step sampling.}
    \label{fig:trajectory}
\end{figure*}

\begin{figure*}[h!]
    \centering
    \begin{subfigure}{0.45\textwidth} 
        \centering
        \includegraphics[width=\textwidth]{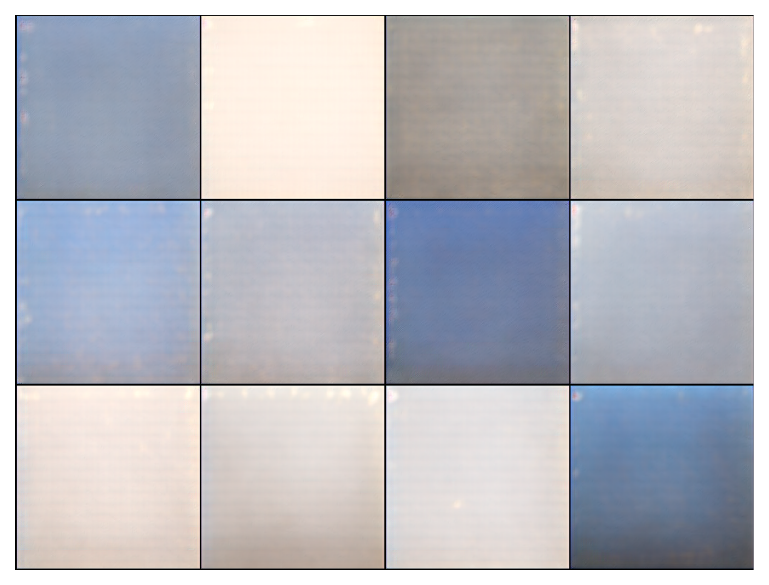}
        \vspace{0pt}
        \caption{Q-diffusion $\left[W4A4\right]$}
    \end{subfigure}
    \begin{subfigure}{0.45\textwidth} 
        \centering
        \includegraphics[width=\textwidth]{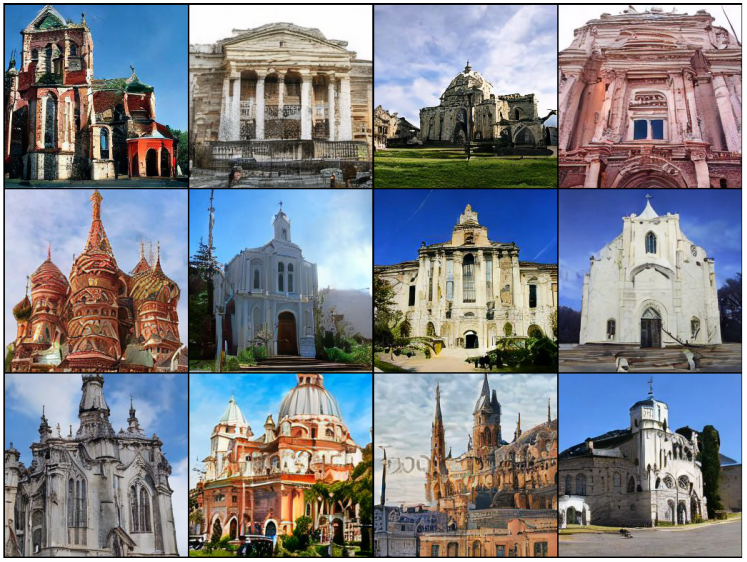}
        \vspace{0pt}
        \caption{SA-QLoRA $\left[W4A4\right]$}
    \end{subfigure}
    \caption{\justifying Generative performance comparison of W4A4 quantized LDM8 models, employing Q-diffusion and our proposed SA-QLoRA, on the LSUN-Church 256$\times$256 dataset with 50-step sampling.}
    \label{fig:trajectory}
\end{figure*}
\begin{figure*}[!t]
    \centering
    \begin{subfigure}{0.45\textwidth} 
        \centering
        \includegraphics[width=\textwidth]{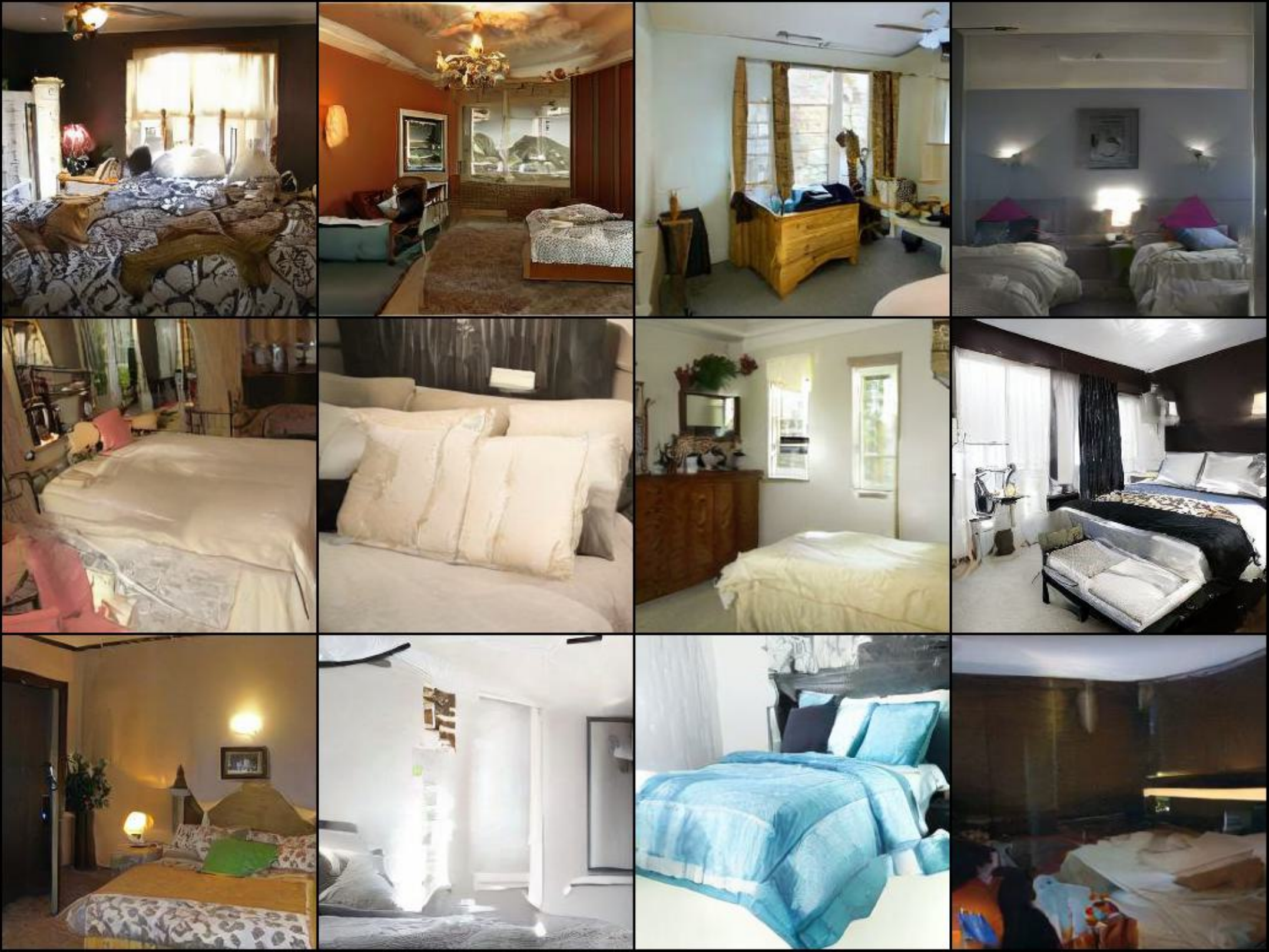}
        \vspace{0pt}
        \caption{FP32}
    \end{subfigure}
    \begin{subfigure}{0.45\textwidth} 
        \centering
        \includegraphics[width=\textwidth]{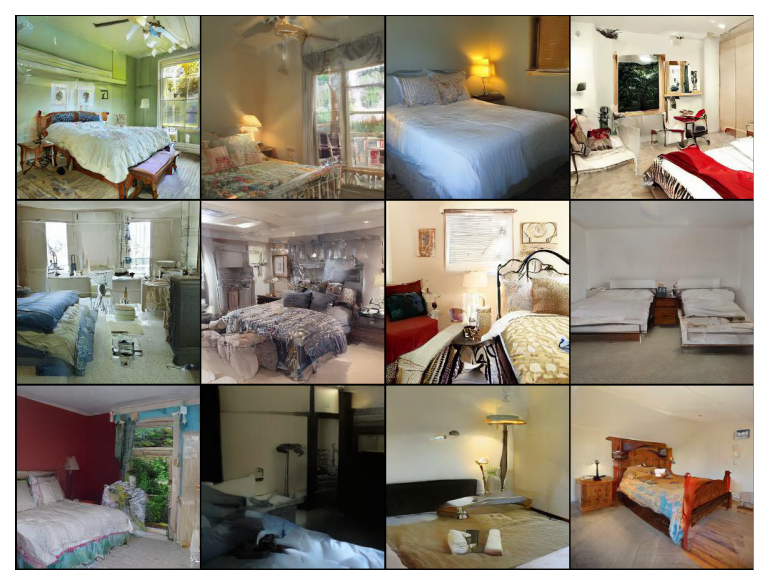}
        \vspace{0pt}
        \caption{SA-QLoRA $\left[W4A8\right]$}
    \end{subfigure}
    \caption{\justifying Comparison of generative performance between the full-precision LDM4 and its W4A8 quantized counterpart, utilizing our proposed SA-QLoRA, on the LSUN-Bedroom 256$\times$256 dataset under 50-step sampling.}
    \label{fig:trajectory}
\end{figure*}

\begin{figure*}[!t]
    \centering
    \begin{subfigure}{0.45\textwidth} 
        \centering
        \includegraphics[width=\textwidth]{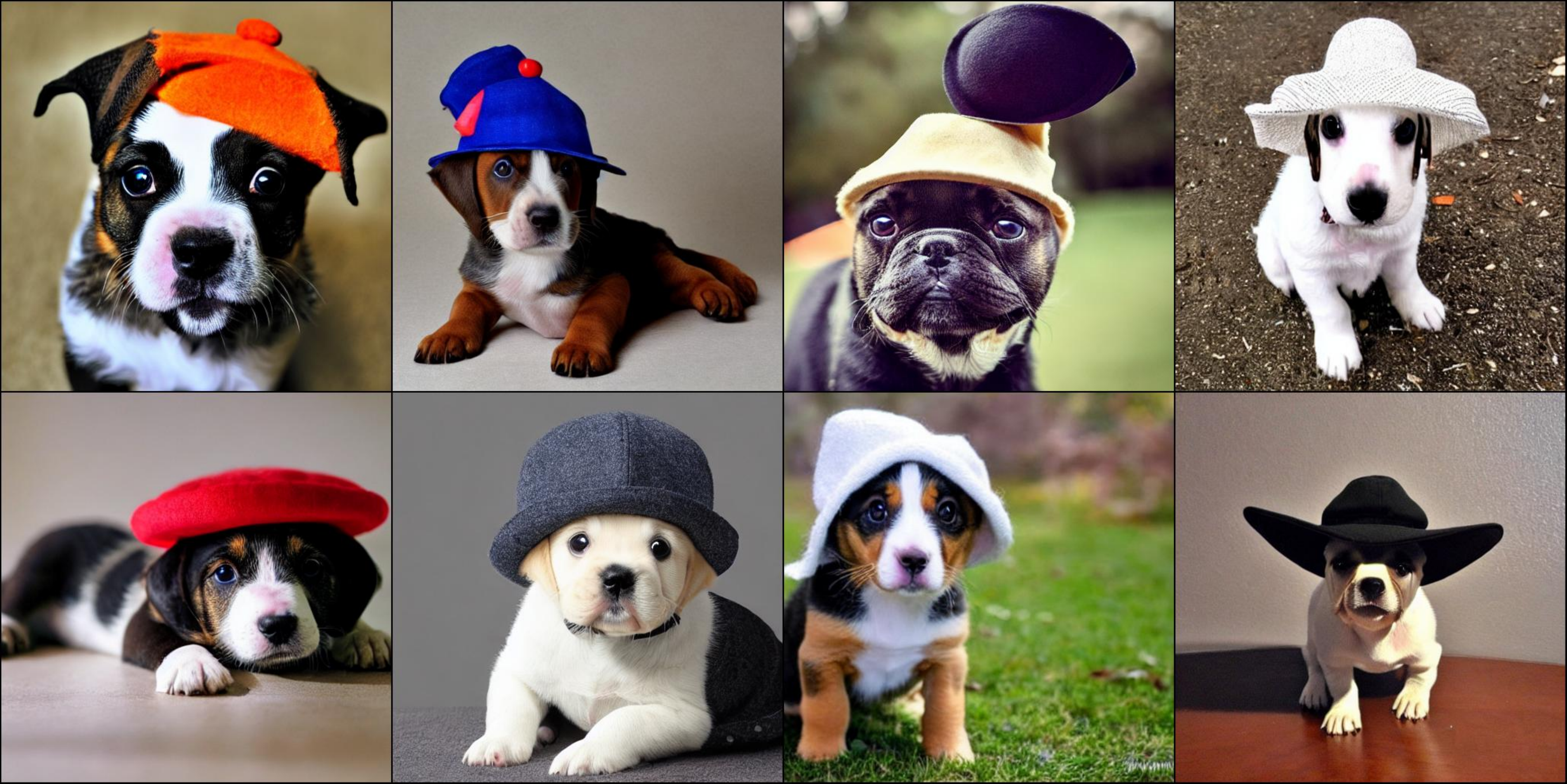}
        \vspace{0pt}
        \caption{Prompt "a puppy wearing a hat"}
    \end{subfigure}
    \begin{subfigure}{0.45\textwidth} 
        \centering
        \includegraphics[width=\textwidth]{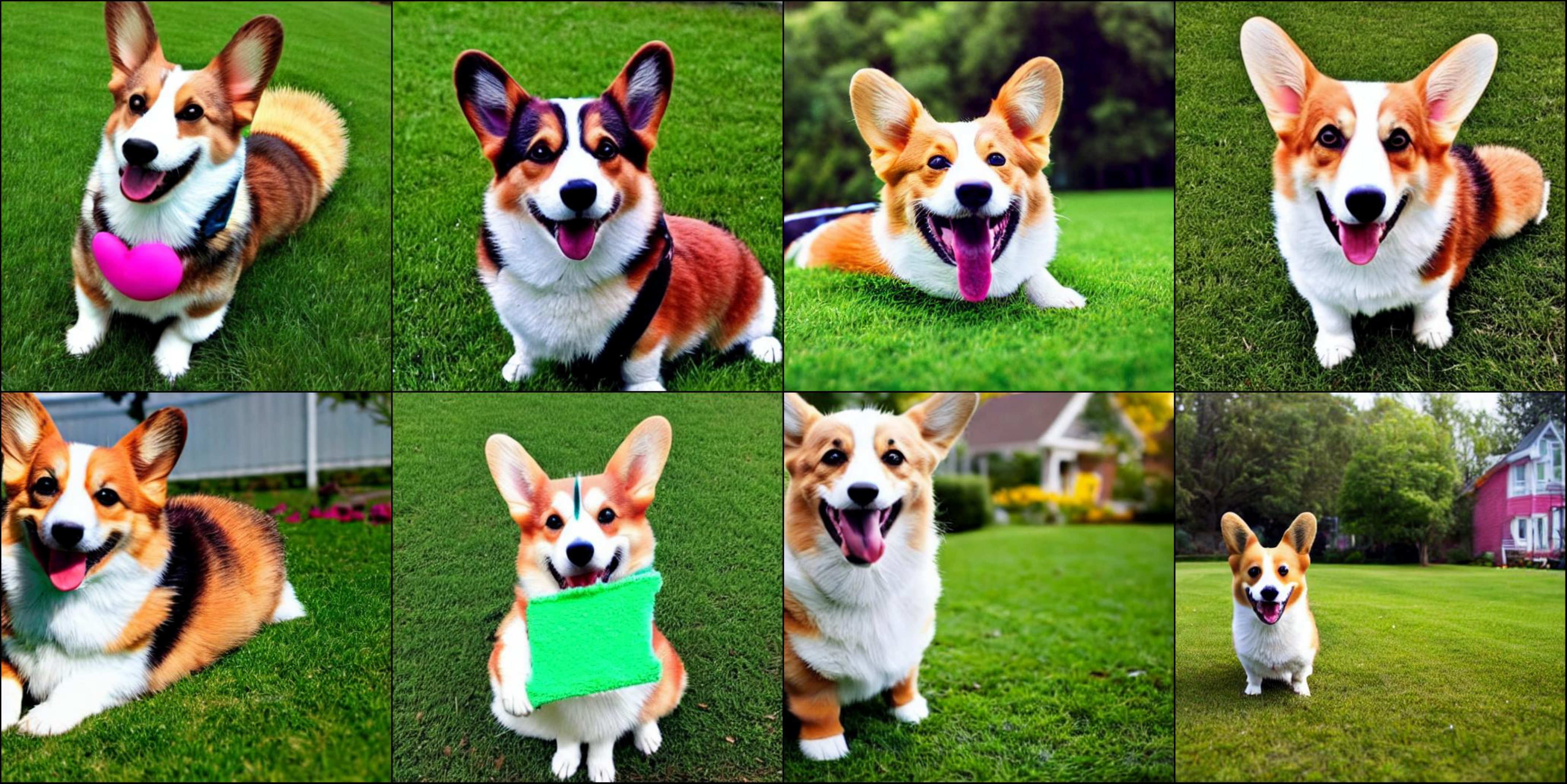}
        \vspace{0pt}
        \caption{Prompt "A Corgi lying on a green lawn, smiling happily."}
    \end{subfigure}
    \caption{\justifying Generation performance of our SA-QLoRA under W8A8 quantization.}
    \label{fig:text2image}
\end{figure*}



\end{document}